\documentclass{egpubl}

\JournalSubmission    %

\usepackage[T1]{fontenc}
\usepackage{dfadobe}  

\usepackage{cite}  %
\BibtexOrBiblatex
\electronicVersion
\PrintedOrElectronic
\ifpdf \usepackage[pdftex]{graphicx} \pdfcompresslevel=9
\else \usepackage[dvips]{graphicx} \fi

\usepackage{egweblnk}

\title[MINERVAS]%
      {MINERVAS: Massive INterior EnviRonments VirtuAl Synthesis}

\author[H. Ren, H. Zhang, J. Zheng, J. Zheng, R. Tang, Y. Huo, H. Bao \& R. Wang]
{\parbox{\textwidth}{\centering Haocheng Ren$^1$\thanks{Equal contributions.}, Hao Zhang$^{1\dagger}$, Jia Zheng$^2$, Jiaxiang Zheng$^2$, Rui Tang$^{2\ddagger}$, Yuchi Huo$^1$, Hujun Bao$^1$ and Rui Wang$^1$\thanks{Corresponding authors: 
Rui Tang, ati@qunhemail.com; 
Rui Wang, rwang@cad.zju.edu.cn.
}
}
\\
{\parbox{\textwidth}{\centering $^1$State Key Lab of CAD\&CG, Zhejiang University,
$^2$Manycore Tech Inc.
}
}
}

\usepackage{amsmath,amsfonts}
\usepackage{algorithmic}
\usepackage{algorithm}
\usepackage{array}
\usepackage[caption=false,font=normalsize,labelfont=sf,textfont=sf]{subfig}
\usepackage{textcomp}
\usepackage{stfloats}
\usepackage{url}
\usepackage{verbatim}
\usepackage{graphicx}
\usepackage{cite}

\usepackage{amssymb}
\usepackage{booktabs}

\usepackage[frozencache,cachedir=.]{minted}
\usepackage{xcolor}
\usepackage{pifont}
\usepackage{enumitem}
\usepackage{multirow}

\newcommand{\cmark}{\ding{51}}
\newcommand{\xmark}{\ding{55}}
\newminted{python}{frame=single,fontsize=\scriptsize}

\newcommand{\PAR}[1]{\smallskip \noindent \textbf{#1}}

\begin{document}

\maketitle
\begin{abstract}
With the rapid development of data-driven techniques, data has played an essential role in various computer vision tasks. Many realistic and synthetic datasets have been proposed to address different problems. However, there are lots of unresolved challenges: (1) the creation of dataset is usually a tedious process with manual annotations, (2) most datasets are only designed for a single specific task, (3) the modification or randomization of the 3D scene is difficult, and (4) the release of commercial 3D data may encounter copyright issue. This paper presents MINERVAS, a Massive INterior EnviRonments VirtuAl Synthesis system, to facilitate the 3D scene modification and the 2D image synthesis for various vision tasks. In particular, we design a programmable pipeline with Domain-Specific Language, allowing users to select scenes from the commercial indoor scene database, synthesize scenes for different tasks with customized rules, and render various types of imagery data, such as color images, geometric structures, semantic labels. Our system eases the difficulty of customizing massive scenes for different tasks and relieves users from manipulating fine-grained scene configurations by providing user-controllable randomness using multi-level samplers. Most importantly, it empowers users to access commercial scene databases with millions of indoor scenes and protects the copyright of core data assets, e.g., 3D CAD models. We demonstrate the validity and flexibility of our system by using our synthesized data to improve the performance on different kinds of computer vision tasks. The project page is at \url{https://coohom.github.io/MINERVAS}.


\end{abstract}  
\section{Introduction}
Recent advances in various computer vision tasks have shown the tremendous capabilities of deep learning algorithms. Meanwhile, datasets have demonstrated their importance in affecting the performance of these data-driven algorithms. Most datasets~\cite{Silberman:ECCV12,chou2020360,song2015sun,dai2017scannet,chang2017matterport3d,xia2018gibson,xiao2013sun3d} are built from the real world, and contain imagery or 3D data (e.g., scanned point clouds or meshes). However, collecting the data and manually annotating is a tedious and time-consuming process.

The creation of synthetic datasets with a lower cost has become an essential endeavor for accomplishing learning tasks to address these issues. Many synthetic datasets have been presented already. Some datasets release images or videos~\cite{mccormac2017scenenet,li2018cgintrinsics, InteriorNet18,Structured3D} for specific vision tasks, others release 3D scenes~\cite{handa2016scenenet,li2020openrooms,song2017semantic,fu20203dfront,roberts2020hypersim} to give users more flexibility to synthesize images or videos. However, given the complexity of 3D scene editing and rendering, modifying 3D scenes to augment datasets for training different vision tasks is a complicated process requiring relevant expertise. Another considerable obstacle to the synthetic dataset is the lack of high-quality 3D assets. Although some commercial companies have massive 3D scene data, they cannot provide it directly to researchers due to copyright restrictions~\cite{song2017semantic} and the huge data storage size (e.g., more than hundreds of Terabytes of 3D CAD models). Hence, a new requirement arises that allowing users to access these massive 3D scenes while protecting the core assets of commercial companies.

In this paper, we introduce MINERVAS, a Massive INterior EnviRonments VirtuAl Synthesis system, to facilitate various vision problems by providing an imagery data synthesis platform. Our system leverages the 3D scenes from the commercial indoor scene database and generates large-scale task-specific synthetic data. To help users customize data, we design a programmable pipeline with Domain-Specific Language (DSL) to control the data synthesis process, such as filtering and editing 3D scenes, defining different desired outputs. Moreover, since the diversity of the data is crucial for learning-based methods, we provide user-controllable samplers at different levels (i.e., scene, entity, and image) to support domain randomization within our system. The proposed system makes the data synthesis for different vision tasks simple and flexible. In addition, the core 3D data assets are not released directly, thereby avoiding potential copyright issues~\cite{song2017semantic}.

We showcase the validity and flexibility of our system with different vision tasks, including room layout estimation, semantic segmentation, and depth estimation. 
Our system can easily augment the training data for these tasks, where it only takes 12 hours to generate 1 million photorealistic images. The experimental results show that performances improve on these tasks.

In summary, the main contributions of our work are: (1) We show the possibility to contribute the power of the high-quality commercial scene database to the community while protecting the copyright of the core assets (e.g., 3D CAD models) for the company. (2) We present a multi-stage programmable pipeline with DSL to generate synthetic images in batch processing. (3) We develop multi-level samplers to improve diversity at scene, entity, and image levels. (4) We show the potential of our system by demonstrating the usefulness of our synthetic data on several vision tasks.

\section{Related Work}

\noindent \textbf{Interior datasets.} Recently, many interior datasets~\cite{armeni20163d,dai2017scannet,chang2017matterport3d,xia2018gibson,replica19arxiv,fisher2012example,handa2016scenenet,song2017semantic,Avetisyan_2019_CVPR,tartanair2020iros,li2020openrooms,fu20203dfuture,fu20203dfront,roberts2020hypersim} have been proposed to facilitate research on various indoor tasks. 
Table~\ref{tab:compare} shows existing datasets, 
which can be broadly divided into two distinct categories: real-world datasets and synthetic datasets.

\begin{table}[t]
    \scriptsize
    \centering
    \setlength{\tabcolsep}{1pt}
    \caption{A comparison of existing indoor scene datasets. Notations: 3D (including point cloud or meshes), O (object detection), U (scene understanding), S (image synthesis), M (structured 3D modeling).}
    \label{tab:compare}
    \begin{tabular}{l|ccccc}
        \toprule
        Real Dataset & Format & \#Scenes & Editable & \#Category & Application \\
        \midrule
        Cornell RGB-D~\cite{NIPS2011_4226} & Image+3D & 52 & \xmark & --- & U O \\
        NYUv2~\cite{Silberman:ECCV12} & Image & 464 & \xmark & 894 & U O \\
        TUM~\cite{sturm12iros} & Image & 47 & \xmark & --- & U O \\
        SUN 3D~\cite{xiao2013sun3d} & Image+3D & 254 & \xmark & --- & U O \\
        B3DO~\cite{janoch2013category} & Image & 75 & \xmark & 50 & U O \\
        RGBDv2~\cite{lai2014unsupervised} & 3D & 17 & \xmark & 51 & U O \\
        SUN RGB-D~\cite{song2015sun} & Image & --- & \xmark & 800 & U O \\
        SceneNN~\cite{hua2016scenenn} & Image+3D & 100 & \xmark & 40 & U O \\
        PiGraphs~\cite{savva2016pigraphs} & Image+3D & 26 & \xmark  & --- & U O \\
        S3DIS~\cite{armeni20163d} & 3D & 265 & \xmark & --- & U O \\
        Stanford 2D-3D-S~\cite{armeni2017joint} & Image+3D & 279 & \xmark  & 13 & U O \\
        Matterport3D~\cite{chang2017matterport3d} & Image+3D & 90 & \xmark & 40 & U O \\
        ScanNet~\cite{dai2017scannet} & Image+3D & 1513 & \xmark & $\approx$1000 & U O \\
        Gibson~\cite{xia2018gibson} & Image+3D & 572 & \xmark & 80 & U O \\
        Replica~\cite{replica19arxiv} & 3D & 18 & \xmark & 88 & U O \\
        360-Indoor~\cite{chou2020360} & Image & --- & \xmark & 37 & U O \\
        HM3D~\cite{ramakrishnan2021hm3d} & Image+3D & 1000 & \xmark & --- & U O \\
        \midrule
        Synthetic Dataset & Format & \#Scenes & Editable & \#Category & Application \\
        \midrule
        Stanford Scenes~\cite{fisher2012example} & 3D & 130 & \cmark & 6 &  U O S \\
        SUNCG~\cite{song2017semantic} & Image+3D& 45622 & \cmark & 84 & U O S \\
        SceneNet~\cite{handa2016scenenet} & 3D & 57 & \cmark & 218 & U O S \\
        AI2-THOR~\cite{kolve2017ai2} & 3D & 120 & \cmark & 102 & U O \\
        SceneNet-RGB-D~\cite{mccormac2017scenenet} & Image & 57 & \cmark & 255 & U O \\
        CGIntrinsics~\cite{li2018cgintrinsics} & Image & 20K+ & \xmark & --- & U O \\
        InteriorNet~\cite{InteriorNet18} & Image & 22M & \xmark & 158 &  U O S \\
        House3D~\cite{wu2018building} & Image & 45K & \xmark & 84 & U O \\
        RobotriX~\cite{garcia2018robotrix} & Image+3D & 16 & \cmark & 38 &  U O S \\
        DeepFurniture~\cite{liu2019furnishing} & Image & --- & \xmark & 11 & U O \\
        Scan2CAD~\cite{Avetisyan_2019_CVPR} & 3D & 1506 & \cmark & 37 & U O S \\
        Structured3D~\cite{Structured3D} & Image+Structure & 3500 & \xmark & 40 &  U O S M \\
        TartanAir~\cite{tartanair2020iros} & Image+3D & 30 & \cmark & --- &  U O S \\
        OpenRooms~\cite{li2020openrooms} & Image+3D & 1506 & \cmark & 24 & U O S \\
        3D-FUTURE~\cite{fu20203dfuture} & Image+3D & 5000 & \cmark & 34 &  U O S \\
        3D-FRONT~\cite{fu20203dfront} & Image+3D & 6813 & \cmark & --- & U O S \\
        Hypersim~\cite{roberts2020hypersim} & Image+3D & 461 & \cmark & --- & U O S \\
        \textbf{MINERVAS (Ours)} & Image+Structure & \textbf{50M} & \cmark & 285 & U O S M ... \\
        \bottomrule
    \end{tabular}
\end{table}

Real-world datasets are usually collected in the real world using RGB-D sensors (such as Microsoft Kinect) and represented as RGB-D sequences~\cite{Silberman:ECCV12,sturm12iros,janoch2013category,song2015sun}, point clouds~\cite{NIPS2011_4226,xiao2013sun3d,lai2014unsupervised,armeni20163d,xia2018gibson} or meshes~\cite{hua2016scenenn, savva2016pigraphs,chang2017matterport3d, armeni2017joint,dai2017scannet,replica19arxiv}.
NYUv2~\cite{Silberman:ECCV12} collected 464 RGB-D sequences and per-pixel annotations in selected 1449 frames.
SUN 3D~\cite{xiao2013sun3d} recovered the full 3D extent of 254 spaces as point clouds instead of a limited set of 2D views.
SceneNN~\cite{hua2016scenenn} reconstructed triangles meshes from more than 100 scenes with fine-grained annotation.
Several works utilize the Matterport system to reconstruct larger scenes as dense meshes~\cite{chang2017matterport3d,armeni2017joint} or point clouds~\cite{armeni20163d,xia2018gibson}. 
However, those datasets provide a limited number of scenes due to the complex capture process.
ScanNet~\cite{dai2017scannet} takes one step further by reconstructing 1513 scenes, which is significantly larger than previous works. It utilizes a portable iPad-based RGB-D camera to capture scenes, which helps to involve more untrained users.
However, the mesh quality and the accuracy of semantic segmentation are not high enough.
On the contrary, Replica~\cite{replica19arxiv} focused on the quality instead of the dataset size. They use a custom-built RGB-D capture rig to obtain 18 high-quality reconstructed scenes. 
Dense meshes, high-resolution HDR textures, and semantic segmentation are also included.
Most recently, HM3D~\cite{ramakrishnan2021hm3d} provides 1000 high-quality reconstructed scenes, but the semantic annotation is not included.
Due to the complexity of data collection and annotation, the sizes of the real-world datasets are usually tiny compared to synthetic datasets, as shown in Table~\ref{tab:compare}. The dataset size constrains its diversity. 
Furthermore, the real-world dataset is usually not editable for augmentation. Besides, the captured or reconstructed data from natural environments usually consists of inherent errors.

With the advantages of large-scale, easy-annotation, and editable, many synthetic datasets are introduced to improve performance on several tasks~\cite{zhang2017physically,li2018cgintrinsics,Structured3D}. SUNCG~\cite{song2017semantic} was a widely used artist-designed synthetic dataset. However, SUNCG is unavailable online due to copyright problems. SceneNet~\cite{handa2016scenenet} and Scan2CAD~\cite{Avetisyan_2019_CVPR} have the capability of modification by sampling objects. However, these datasets are lack realism and semantic diversity. Recent synthetic datasets only release a few 3D data~\cite{fisher2012example,handa2016scenenet,kolve2017ai2,garcia2018robotrix,tartanair2020iros,roberts2020hypersim} or imagery data~\cite{InteriorNet18,mccormac2017scenenet,li2018cgintrinsics,wu2018building,liu2019furnishing}. Most recently, OpenRooms~\cite{li2020openrooms} release an interior synthetic dataset built upon publicly available real scanned datasets, but the diversity of scenes and objects are not as good as artist-created datasets such as SUNCG. We build upon the professional artist-created interior designs, like InteriorNet~\cite{InteriorNet18} and Structured3D~\cite{Structured3D}, but our system allows users to manipulate the scenes and synthesize task-specific data using DSL. Furthermore, our database has the largest number of scenes among existing datasets as shown in Table~\ref{tab:compare}.

Though the works discussed above contribute many interior datasets, few works focus on the copyright protection of 3D assets. A classical and useful way to protect 3D assets is remote rendering~\cite{koller2004protected, koller2005protecting}. Several strategies like adding perturbations are also proposed~\cite{koller2005protecting} to prevent the accurate reconstruction of 3D models.
Recently, MeshChain~\cite{park2021meshchain} utilizes blockchain technology to protect 3D models and enable data source authentication. 
Our system can protect 3D assets via remote rendering.

\smallskip \noindent \textbf{Automatic scene generation.} Applications in computer graphics and computer vision have inspired many works on automatic scene editing, especially 3D scene generation~\cite{merrell2011interactive,yu2015clutterpalette,kan2018automatic,ritchie2019fast,wang2019planit,li2019grains,zhang2020deep,zhang2021fast,zhang2021geometry}. These methods assist users by suggesting novel furniture layouts under given constraints. K{\'a}n and Kaufman~\cite{kan2018automatic} proposed a new automatic furniture arrangement method using greedy cost optimization and achieved good interactive speed. Recently, several works have shown that probabilistic programming language can be used to define distribution to generate scenes, such as WebPPL~\cite{goodman2014design}, Picture~\cite{kulkarni2015picture}, and Scenic~\cite{fremont2019scenic}. More recently, Jiang et al.~\cite{jiang2018configurable} proposed a configurable approach to generate synthetic scenes, which is in some sense the most closely related to our work. Nevertheless, their work focuses on scene synthesis rather than a system that enables batch processing and user-controllable 3D scene editing from a large-scale database.

\begin{figure*}[t]
    \centering
    \includegraphics[width=\linewidth]{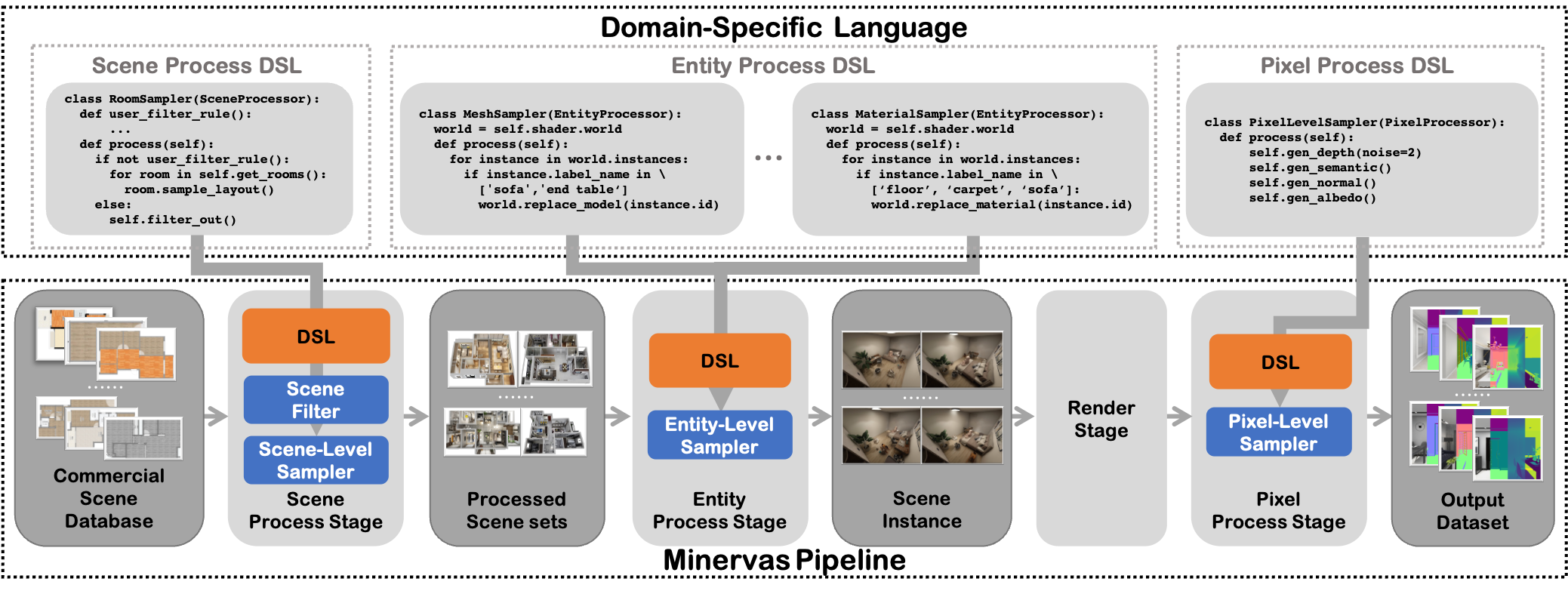}
    \caption{Overview of MINERVAS system. Our system has four stages to synthesis scenes and render images, DSLs are injected into stages for customized user control.}
    \label{fig:pipeline}
\end{figure*}

\smallskip \noindent \textbf{Programmable image synthesis.} Several works~\cite{johnson2016malmo,kempka2016vizdoom,richter2016playing} have involved adapting the real-time rendering techniques used in computer games to generate the image and video datasets. Several simulators with open-source API~\cite{dosovitskiy2017carla,bondi2018airsim,xia2018gibson,savva2019habitat} have been proposed for various tasks. Habitat-Sim~\cite{savva2019habitat} provided an environment for training embodied AI agents, especially for navigation tasks. However, our system aims to generate imagery datasets for different vision tasks. Furthermore, it is mainly built on real-world datasets (e.g., Replica~\cite{replica19arxiv}, Matterport3D~\cite{chang2017matterport3d}, Gibson~\cite{xia2018gibson}), which makes it hard to edit scenes like synthetic data, as we discussed above.
These methods efficiently simulate a single scene, but they are not designed for easy scene modification and data generation in batch processing.

Most recently, several systems~\cite{denninger2019blenderproc, morrical2021nvisii, gan2021tdw, eftekhar2021omnidata, borkman2021unity, Greff_2022_CVPR} are proposed to generate synthetic dataset from collections of 3D assets. Though the overall goal is similar, our system differs from them in several aspects.
BlenderProc~\cite{denninger2019blenderproc} provides a configurable system that can generate photorealistic large-scale imagery datasets from public 3D scene databases, but it has limited scriptable interfaces. NviSII~\cite{morrical2021nvisii}, ThreeDWorld~\cite{gan2021tdw} and Kubric~\cite{Greff_2022_CVPR} uses python API to address this problem. It enables fine-grained control of the scene, but it brings a tedious coding process lacking some high-level built-in functions for randomization. 
Our system provides a user-friendly DSL, considering both flexibilities with the Python programming language and ease of use with built-in randomization support. 
Our multi-stage pipeline design also allows users to modify the parts they are interested in and prevent writing the complete data generation code.

Aside from the system design, 3D assets in our system also differ from them.
Omnidata~\cite{eftekhar2021omnidata} generates imagery datasets using 3D scan data from the real world. Since it uses the 3D scans database, scenes are not editable thus lack further domain randomization (e.g., novel furniture arrangement) like our artist-created synthetic scenes. Unity Perception~\cite{borkman2021unity} also provides a flexible interface for users. However, it uses the Unity Asset Store as the database, the quality of assets is not ensured,
and manually annotating assets is still required. 
The concurrent work Kubric~\cite{Greff_2022_CVPR} mostly relies on existing public 3D assets.
Instead, there is a high-quality large-scale commercial scene database behind our system for users to access as an online service. 
Moreover, some works~\cite{borkman2021unity,gan2021tdw} render dataset based on rasterization for fast rendering, while others~\cite{denninger2019blenderproc,morrical2021nvisii, Greff_2022_CVPR} rely on a path tracer for photorealistic rendering. Our system provides both types of renderers for users to trade-off efficiency and realism.

\section{System Overview}

MINERVAS system aims to provide a platform for users to easily and efficiently synthesize the large-scale, high-quality interior imagery dataset. To this end, we introduce a novel programmable dataset synthesis pipeline, as shown in Figure~\ref{fig:pipeline}.

\subsection{Design Goals}\label{sec:design_goals}

Our system was guided by the following goals:
\begin{itemize}
    \item \PAR{High-quality data:}
    The quality of data is important to deep learning. The higher level of realism could reduce the domain gap~\cite{zhang2017physically} between the synthetic dataset and the real world. The creation of a high-quality synthetic imagery dataset requires both elaborately designed 3D assets and state-of-the-art photo-realistic rendering techniques.

    \item \PAR{Large-scale data:}
More data usually brings better results. One important goal of our system is to provide the largest number of 3D interior scenes among existing synthetic interior datasets.

    \item \PAR{Diverse data:}
    Domain randomization is a common technique to boost the model performance in the real case~\cite{tobin2017domain}. To increase the diversity of the synthesized data, we need to provide both the diverse 3D scenes in the database and the ability to easily augment these 3D scenes in our system.

    \item \PAR{Full user control:} 
    As our system aims to facilitate a broad spectrum of vision tasks, users need to control the system at finer granularity to meet their requirements.

    \item \PAR{Ease of use:} 
    With the guarantee of full user control, we also expect our system to provide a user-friendly control interface, reducing the time users are familiar with the system.
    
    \item \PAR{Fast synthesis:}
    Rendering a large-scale synthetic dataset is a tedious and time-consuming process~\cite{zhang2017physically}. We expect our system can generate synthetic datasets as fast as possible while ensuring the quality of the data.
    
\end{itemize}

\subsection{Pipeline Overview}

Our system pipeline consists of four stages, as shown in Figure~\ref{fig:pipeline}. In the Scene Process Stage, users first select a set of scenes from the commercial scene database and rearrange the furniture layout for scene-level randomization. Then, in the Entity Process Stage, users could manipulate a single object for entity-level randomization. Next, 2D imagery data is rendered in the Render Stage. Finally, in the Pixel Process Stage, the users could apply pixel-wise modifications to the 2D renderings, such as simulating sensor noises.

To improve the diversity of the synthesized dataset, we introduce a randomization mechanism in our system called the sampler. We design multi-level samplers for different stages according to the types of data being processed. Users could control the data synthesis process in all stages of the pipeline via DSL at different levels of granularity control while maintaining ease of use.

\begin{figure*}[t]
    \centering
    \includegraphics[width=\linewidth]{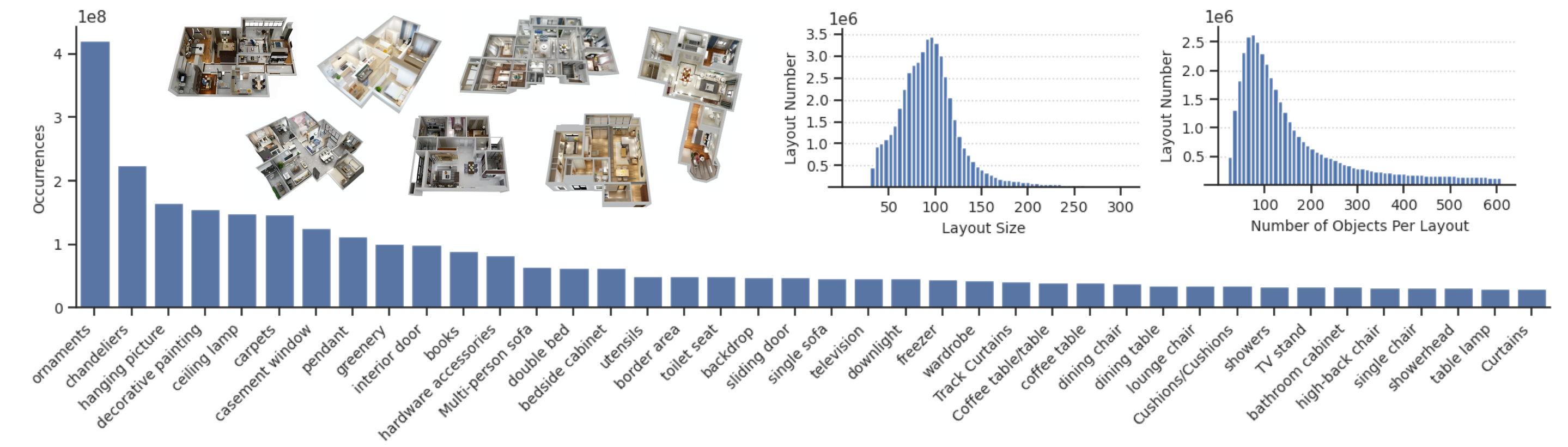}
    \caption{Statistics of the scene database. We show the 40 most frequent categories in the database, the distribution of room size ($\textrm{m}^2$) per layout, the distribution of the number of objects per layout.}
    \label{fig:statistics}
\end{figure*}

\subsection{Key Design Decisions}

To meet the goals in Section~\ref{sec:design_goals}, the following features are considered when designing our system.

\PAR{Large-scale high-quality commercial scene database.} The quality of 3D assets largely affects the quality of the output data. To achieve the best quality and largest number of 3D scenes as we can, we leverage the online commercial 3D scene database from the interior design company
. The underlying database consists of 50 million house designs created by professional interior designers. The dataset has diverse room layouts, furniture arrangements, and lighting setups. Figure~\ref{fig:statistics} shows the statistics of the database. The house layout encompasses a wide range of realistic residential spaces. The database also includes 1 million Computer-Aided Design (CAD) furniture models. These models have been categorized into about 300 categories. We also provide category mappings to commonly-used label sets, such as NYUv2 40 label set~\cite{Silberman:ECCV12}. Each model has physically-based spatially-varying material. This commercial 3D database empowers the ability of our system to generate diverse data by sampling furniture, material, and lighting.

\PAR{Modular system architecture.} By examining the entire process of generating one synthetic image, we found it mainly involves three types of data, i.e., 3D scenes in the database, objects in scenes, and rendered images. For each type of data, processing operations should be different and specific computations need to be customized for different vision tasks. Therefore, we divided the pipeline into four stages. Each stage processes a single type of data in batch. Modulation design improves the flexibility and throughput of our system.

\PAR{Diverse synthesis results.} The diversity and variety of data for training are important for data-driven algorithms. Large-scale datasets with diverse samples are essential to ensure that the algorithm performs adequately in unusual circumstances or challenging cases. In some cases, the users may also need to augment data by varying specific parameters in a scene. Our system enables user-controllable randomness through samplers at different granularity levels of the scenes, entities, and images to improve the data diversity (Section~\ref{sec:scene_synth}).

\PAR{User customization.} Different tasks have different requirements on synthetic datasets. Flexible user customization is an essential feature of our system to meet the needs of different vision tasks. Specifically, we employ a DSL to plug in custom computation in different stages, thus letting users control the process of image generation for their tasks. Additionally, users can utilize the DSL to modify the built-in samplers, letting the randomization be customized (Section~\ref{sec:dsl}). In addition to customization using DSL, users who do not need complex filtering or randomization procedure can use the GUI of our system to customize their generation process while reducing the time of learning the DSL. Compared with DSL's full user customization, the GUI scarifies some freedom but makes our system easier to use.

\PAR{Cloud-based rendering.} As the data synthesis is a compute-intensive task, we deployed our system on a cloud-based cluster with more than 5000 computation nodes, each one composed of the Intel Xeon CPU with 64 cores. Our system takes about 12 hours to generate a large-scale imagery dataset with 1 million views based on the powerful cluster.
In some cases, users want to balance rendering costs and time. For example, some users do not need fast computation on the powerful cluster. 
We also provide the option to render on the local machine to facilitate this.

\PAR{Online service.} Our new programmable system pipeline allows us to implement it as an online service, which can further reduce the tedious process of deploying the system on local machines. Note that implementing our system as an online service can also protect the copyright of these commercial assets~\cite{koller2005protecting}, because the users cannot directly download the core assets (e.g., 3D CAD models) of the database via our online service.
The researchers could re-distribute the data generated on our platform, and we promise the generated data is free from copyright issues. As far as we know, it is the first time to introduce such a massive scale commercial interior 3D scene database to the research community without copyright issues.

\section{Random Scene Synthesis}
\label{sec:scene_synth}

One advantage of synthetic datasets over real-world datasets is that users can freely edit each part of the scenes to meet their dataset generation needs and enable domain randomization. However, editing scenes usually requires relevant expertise, and it is usually tedious to edit massive numbers of scenes. Also, for large-scale dataset synthesis, it is impractical to configure the parameters for each scene manually.

In our system, we introduce a randomness mechanism called sampler to ease the difficulty of editing the scene and improve the diversity of the synthesized datasets. According to the types of data being processed in the different stages of the pipeline (Figure~\ref{fig:pipeline}), we design multi-level samplers to automatically perform scene-level furniture arrangement, entity-level attributes modification, and pixel-level processing.

\subsection{Scene Process Stage}

In the Scene Process Stage, users first select scenes from the commercial scene database according to user-provided conditions, e.g., area, the number of rooms, and room types. To support the mega-scale scene database, we use MongoDB
as the underlying database program. We store all structured metadata of the scene in the database, such as scene graph and scene attributes (e.g., room size, the amount of furniture in the scene).

\begin{figure*}[t]
    \centering
    \footnotesize
    \setlength{\tabcolsep}{1pt}
    \begin{tabular}{cccc}
        
        \includegraphics[width=0.24\linewidth]{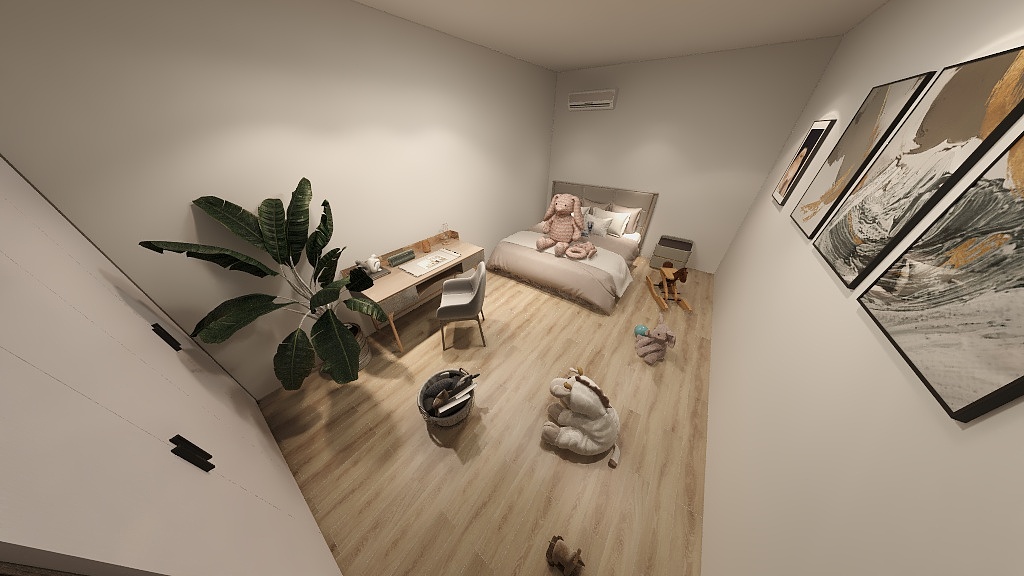} &
        \includegraphics[width=0.24\linewidth]{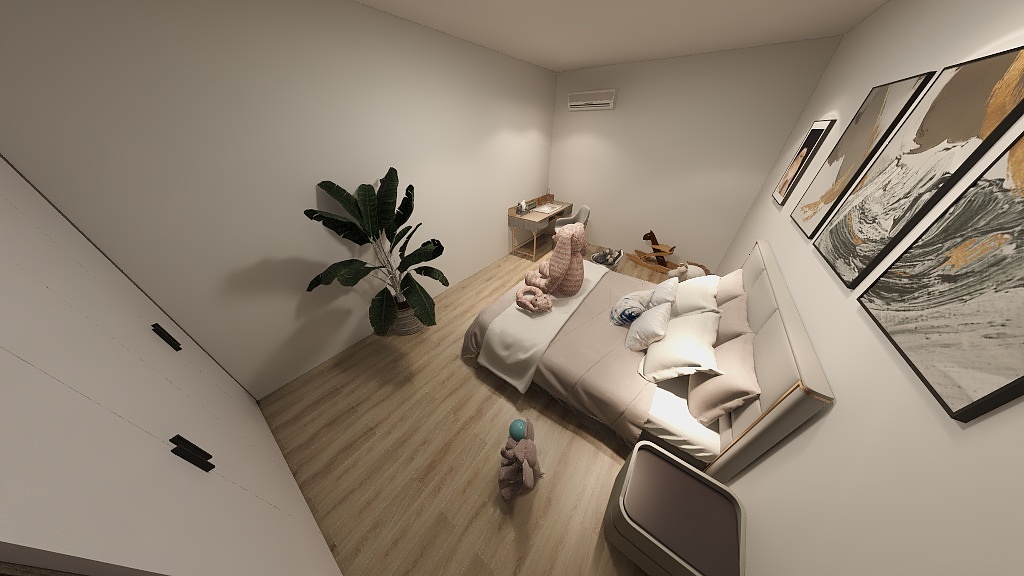} &
        \includegraphics[width=0.24\linewidth]{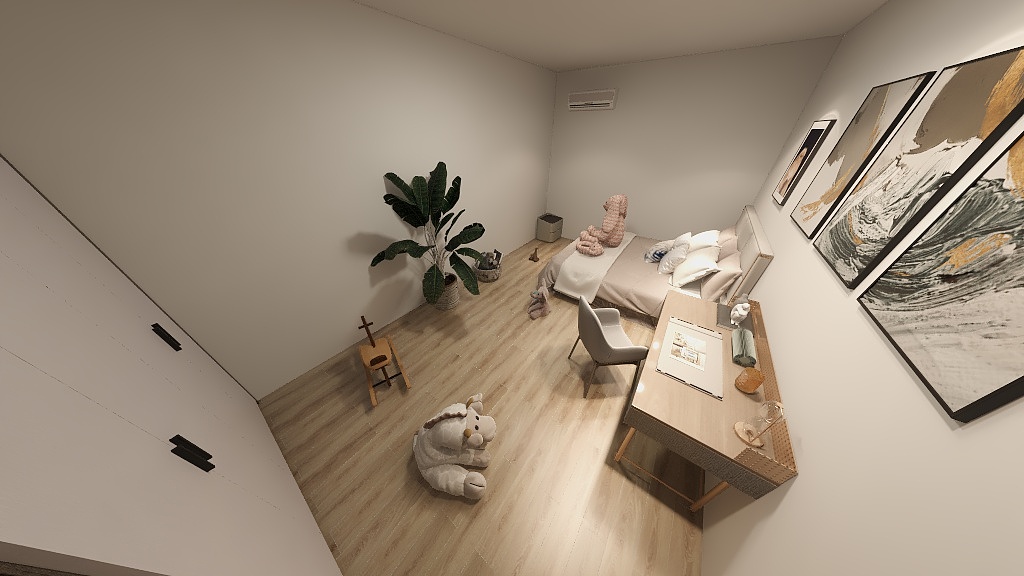} &
        \includegraphics[width=0.24\linewidth]{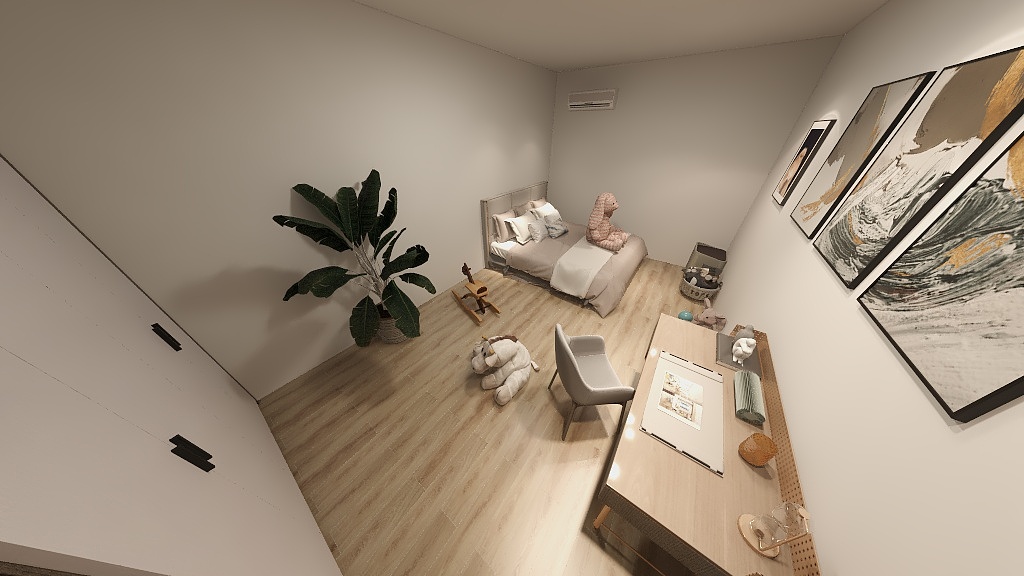} \\
        \multicolumn{4}{c}{(a)} \\
        \includegraphics[width=0.24\textwidth]{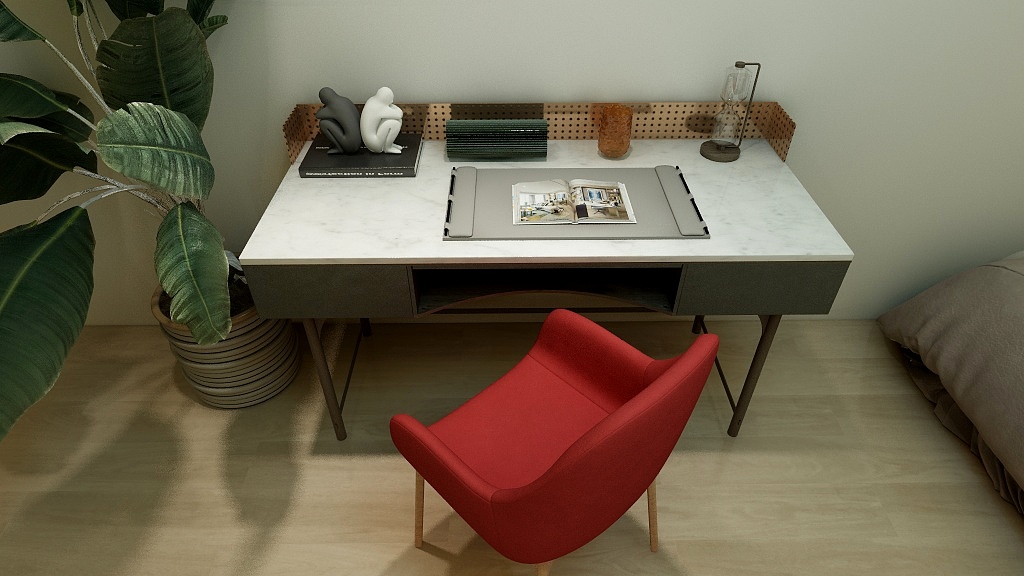} &
        \includegraphics[width=0.24\textwidth]{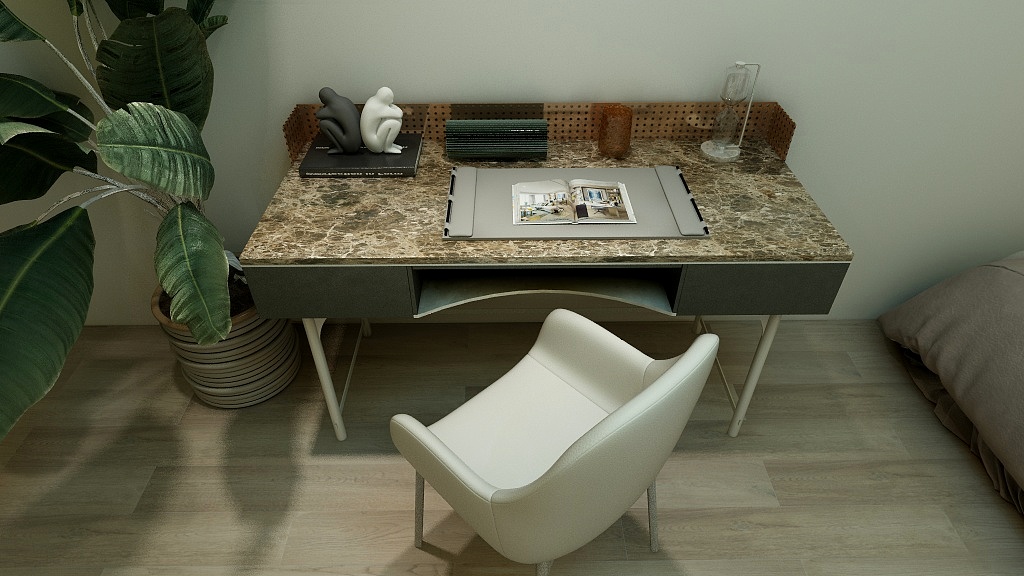} &
        \includegraphics[width=0.24\textwidth]{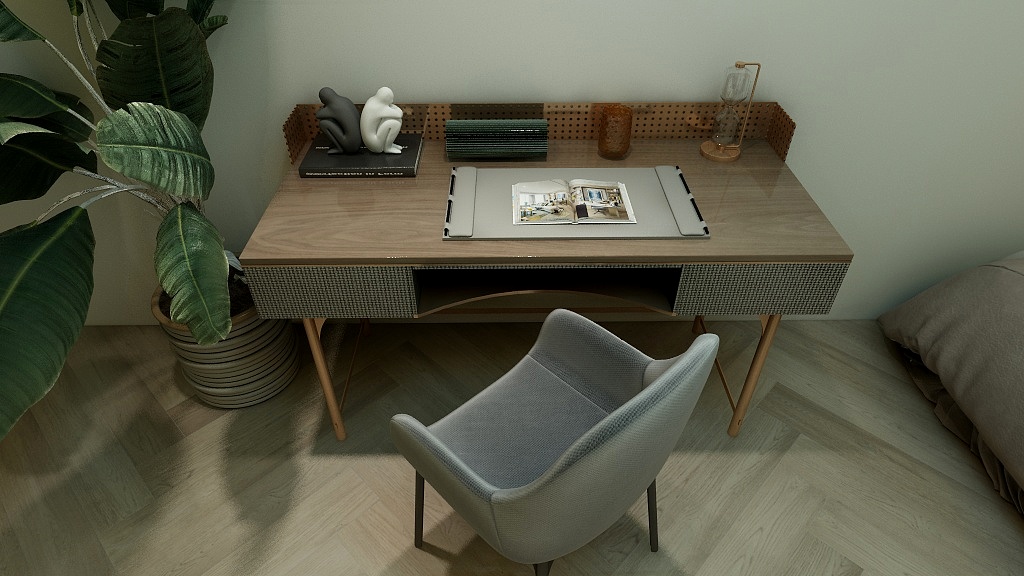} &
        \includegraphics[width=0.24\textwidth]{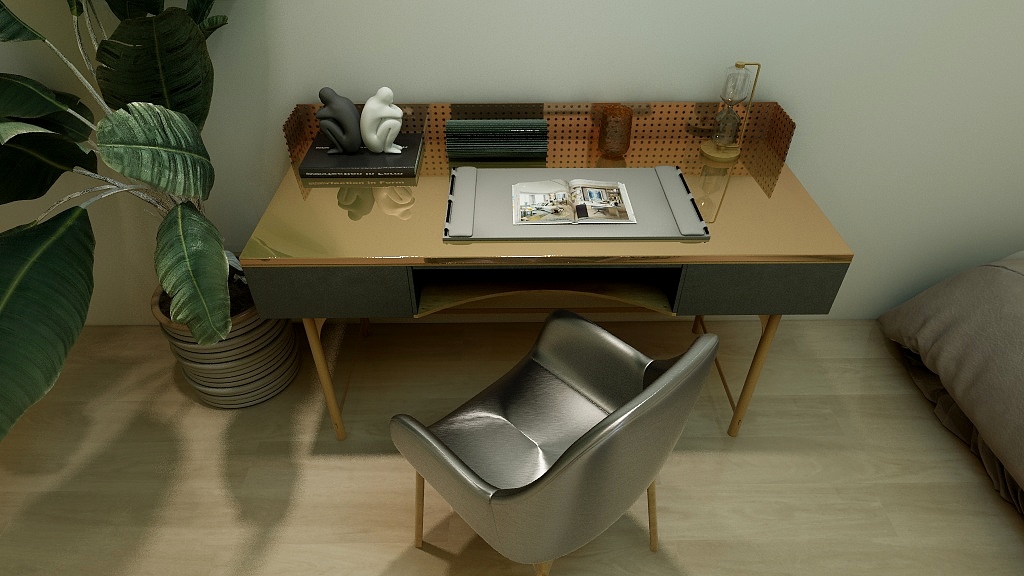} \\
        \multicolumn{4}{c}{(b)} \\
        \includegraphics[width=0.24\textwidth]{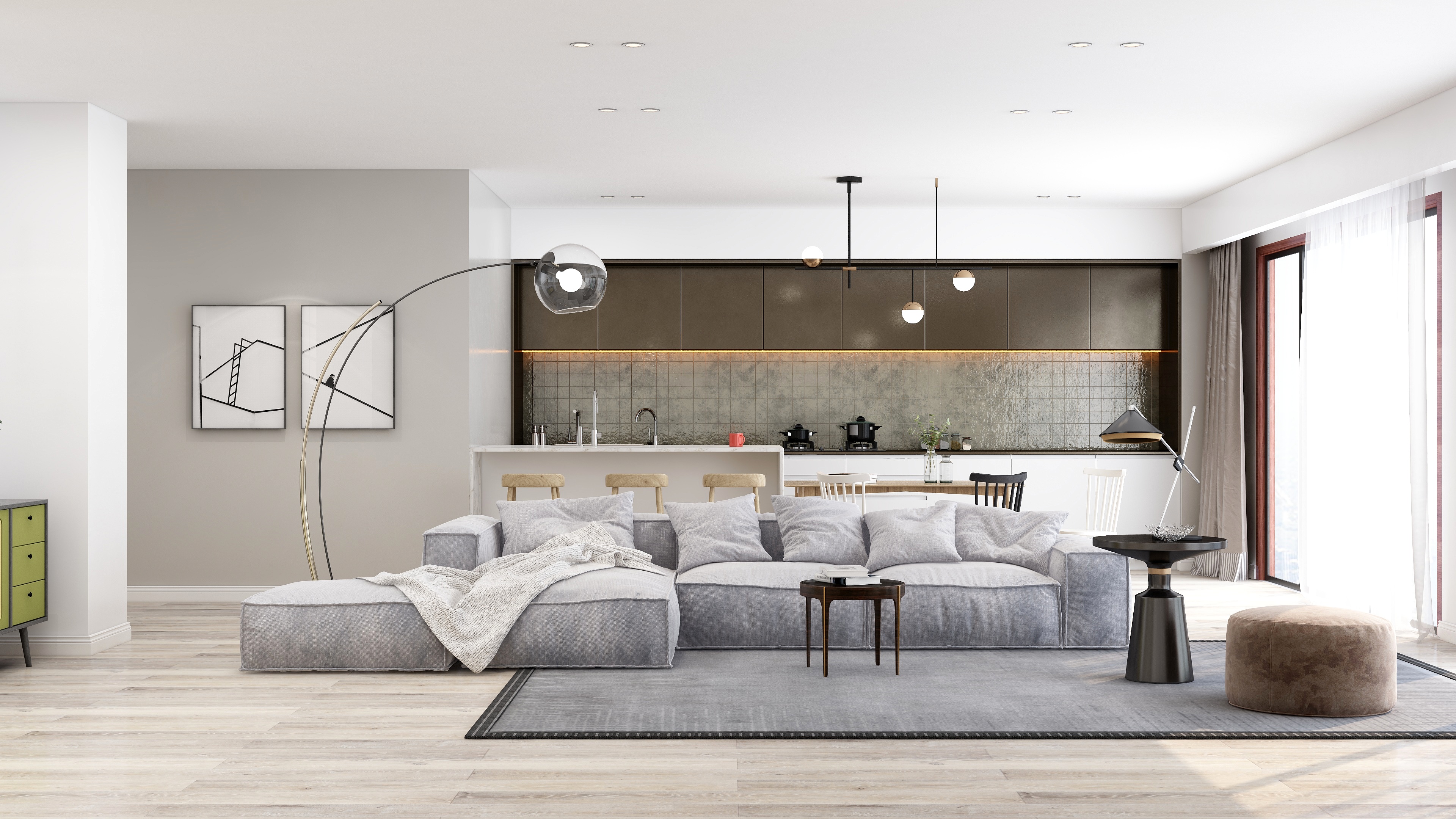} &
        \includegraphics[width=0.24\textwidth]{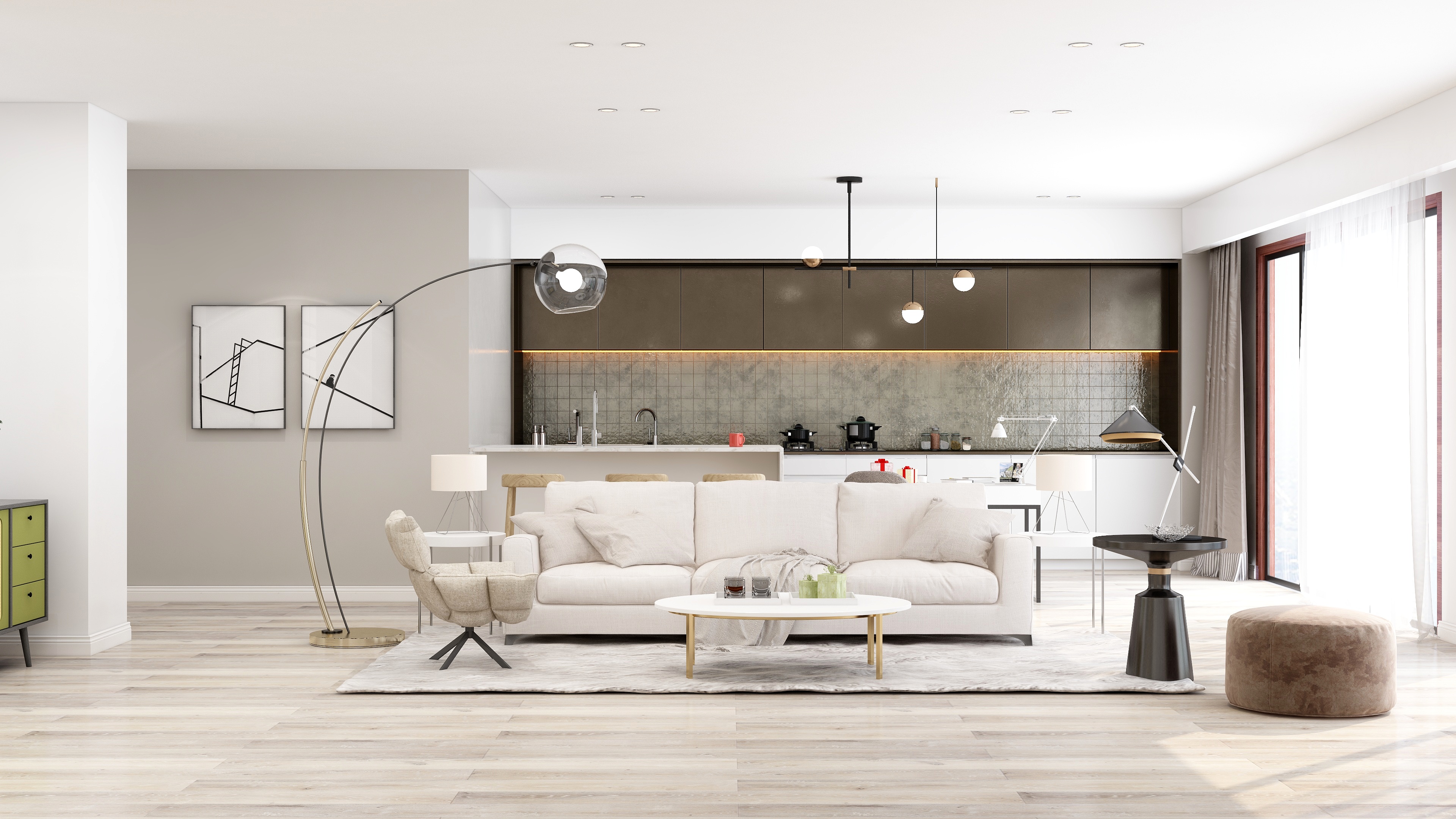} &
        \includegraphics[width=0.24\textwidth]{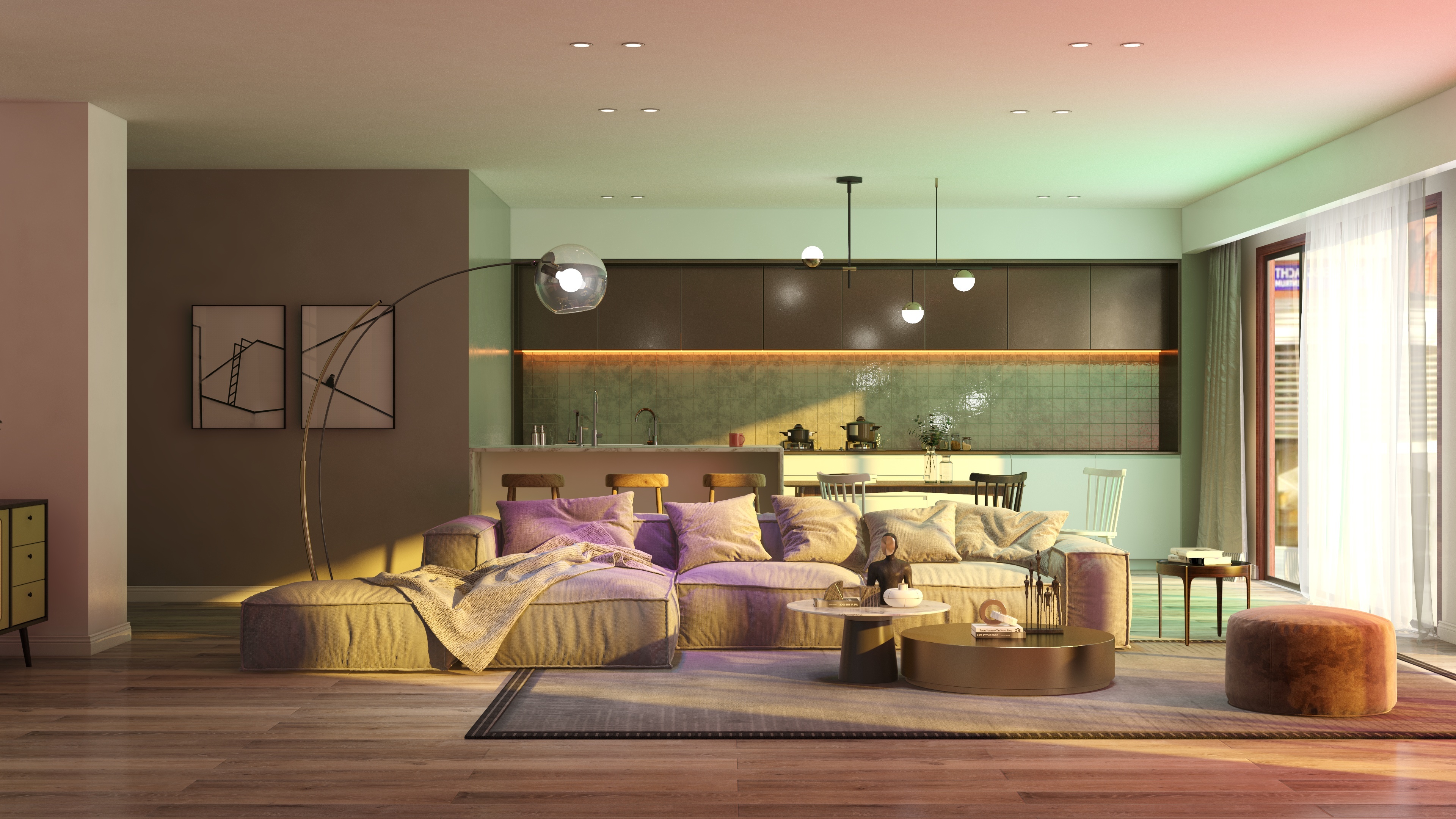} &
        \includegraphics[width=0.24\textwidth]{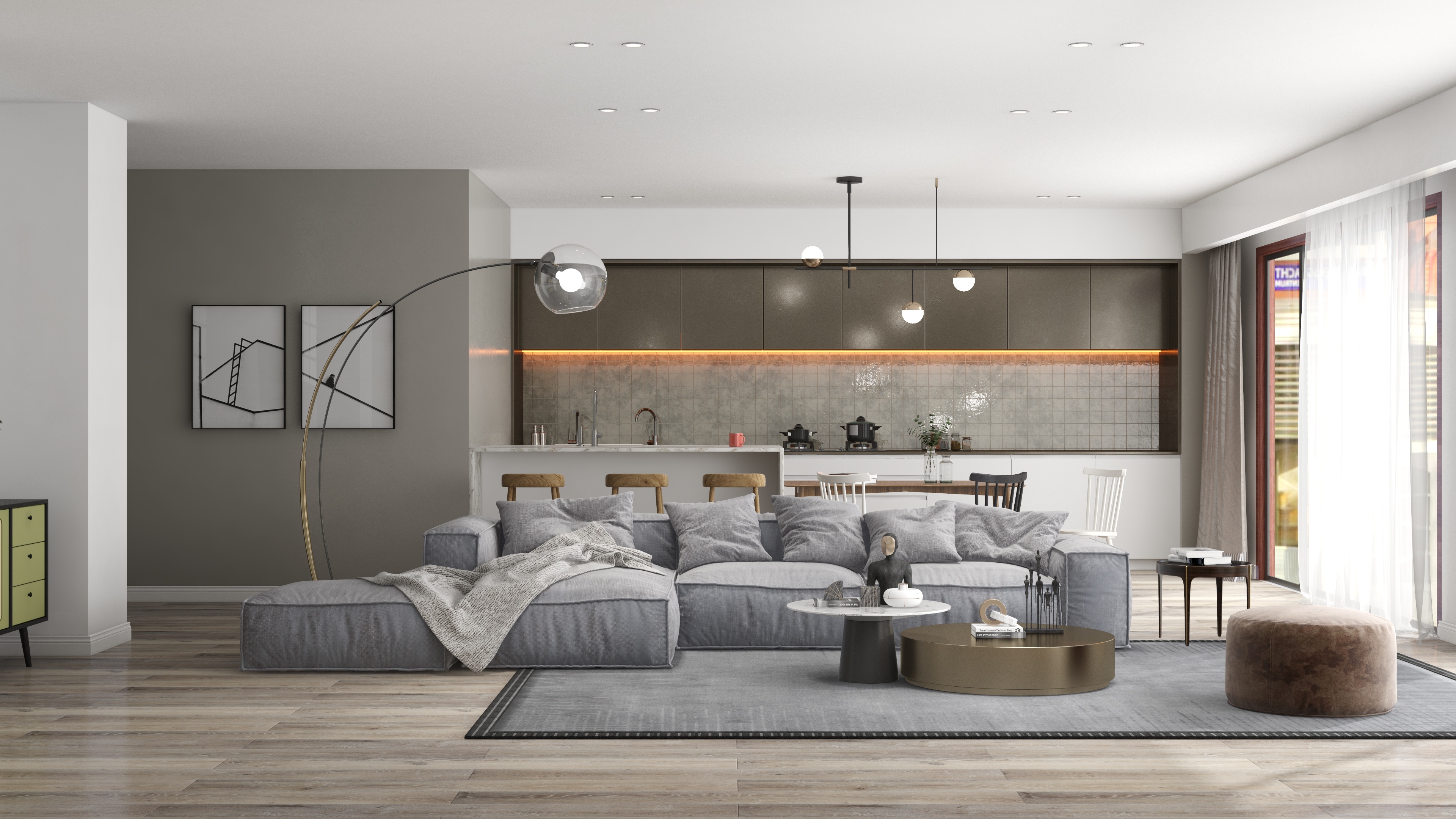} \\
        \multicolumn{2}{c}{(c)} & \multicolumn{2}{c}{(d)} \\
    \end{tabular}
    \caption{Results generated by different samplers. (a) is the results of the scene-level sampler. (b--d) are the results of the entity-level samplers. (b) is the results of randomizing materials. (c) is the results of replacing meshes. (d) is the results under different lighting.}
    \label{fig:property}
\end{figure*}

Given a selected scene, we provide a scene-level sampler to generate novel furniture arrangements. Specifically, we adopt the approach~\cite{kan2018automatic} by taking into account existing furniture and cost functions on the clearance, circulation, group relationships, alignment, distribution, and rhythm. We iteratively generate new arrangements by randomly moving each furniture instance, i.e., changing its positions and orientations. After each iteration, a cost value is calculated to accept or reject the generated layout. The number of iterations is usually determined by the amount of furniture in the scene. Users could easily generate various reasonable furniture arrangements with this scene-level sampler. Figure~\ref{fig:property} (a) shows several layouts generated by the scene-level sampler.

The processed scenes are then sent to the Entity Process Stage for further processing.

\subsection{Entity Process Stage}

The Entity Process Stage is designed for processing each object in the scene. 

We provide some entity-level samplers to randomize attributes of each entity, including furniture (e.g., CAD model, material, transformation), light (e.g., intensity, color), and camera (e.g., camera model, transformation).
Depending on the characteristics of each attribute, we use different types of distributions for different attributes. For example, we utilize uniform, Gaussian distributions to describe continuous attributes (e.g., position, light), discrete distribution for discrete attributes (e.g., material), and learning-based distributions for other attributes (e.g., CAD model). Figure~\ref{fig:property} (b-d) shows the results of sampling different attributes using the entity-level sampler.

Here, the common attributes are shown as follows:

\begin{itemize}
    \item \textbf{Camera.} We support various camera models (e.g., orthographic, perspective, and panoramic camera models) and camera parameters (e.g., field of view, image resolution).

    \item \textbf{Trajectory.} We explicitly define trajectory as an attribute of the movable entity, e.g., camera. To randomly generate roaming trajectories in 3D space, we implement a two-physical-bodies trajectory simulation method~\cite{handa2016scenenet}, i.e., taking a uniform sampler to generate random forces, creating a series of random pivots of the trajectory. Users could also generate the handcrafted trajectory by specifying key points.

    \item \textbf{Light.} Users could control the intensity and color temperature of each light. We assume these attributes follow the uniform distribution. We also support day and night lighting modes, which restrict the range of intensity and color to produce natural lighting.

    \item \textbf{Material}. Given a material category, the sampler uniformly samples materials of that category from a built-in table.

    \item \textbf{CAD model.} In practice, uniform sampling CAD models may not produce good results. Instead, we use an internal similarity map among all CAD models to guide the sampler. More specifically, based on the preview image of CAD model, we use VGGNet~\cite{Simonyan15} to construct a feature vector for each CAD model and perform similarity retrieval based on feature vectors according to their Euclidean metric. When replacing the CAD model, we place a new CAD model in the original place and make sure the object touches the floor or a supported object.

    \item \textbf{Transformation.} The sampler could sample rotation and translation within a given range from a normal or a uniform distribution.
    To avoid intersections, users can enable collision detection using rejection sampling in the sampler.
\end{itemize}

Furthermore, this stage supports to export the non-imagery data (e.g., 3D structures~\cite{Structured3D}) to fulfill specific task requirements such as room layout estimation. The non-imagery data could include any meta information except the raw data of 3D mesh and material.

The generated scenes are then sent to the Render Stage.

\subsection{Render Stage}

In Render Stage, the system uses the generated scenes to generate 2D renderings. Our system provides both rasterization-based and path-tracing-based renderers to balance efficiency and realism. This stage is also configurable, e.g., the number of samples, the number of light bounces. The renderer supports different kinds of imagery data, e.g., color, depth, normal, semantic, instance. The rendered pixel-wise ground truths could facilitate various indoor scene understanding tasks. After this stage, the 2D renderings will be sent to the Pixel Process Stage.

\subsection{Pixel Process Stage}

The Pixel Process Stage allows users to apply pixel-wise processing operations on the imagery data.

\begin{figure}[t]
    \centering
    \footnotesize
    \setlength{\tabcolsep}{1pt}
    \begin{tabular}{cccc}
        \includegraphics[width=0.24\linewidth]{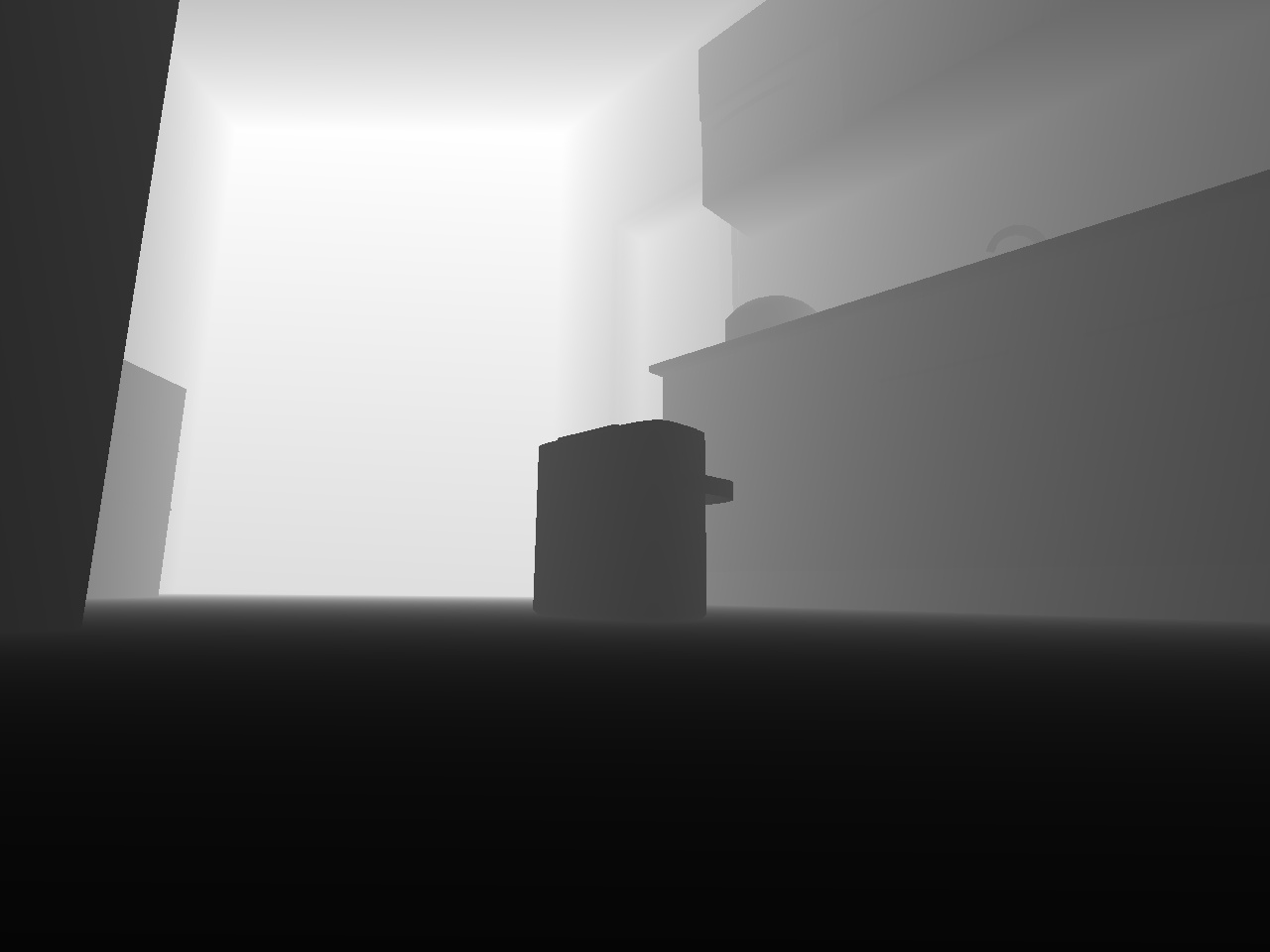} &
        \includegraphics[width=0.24\linewidth]{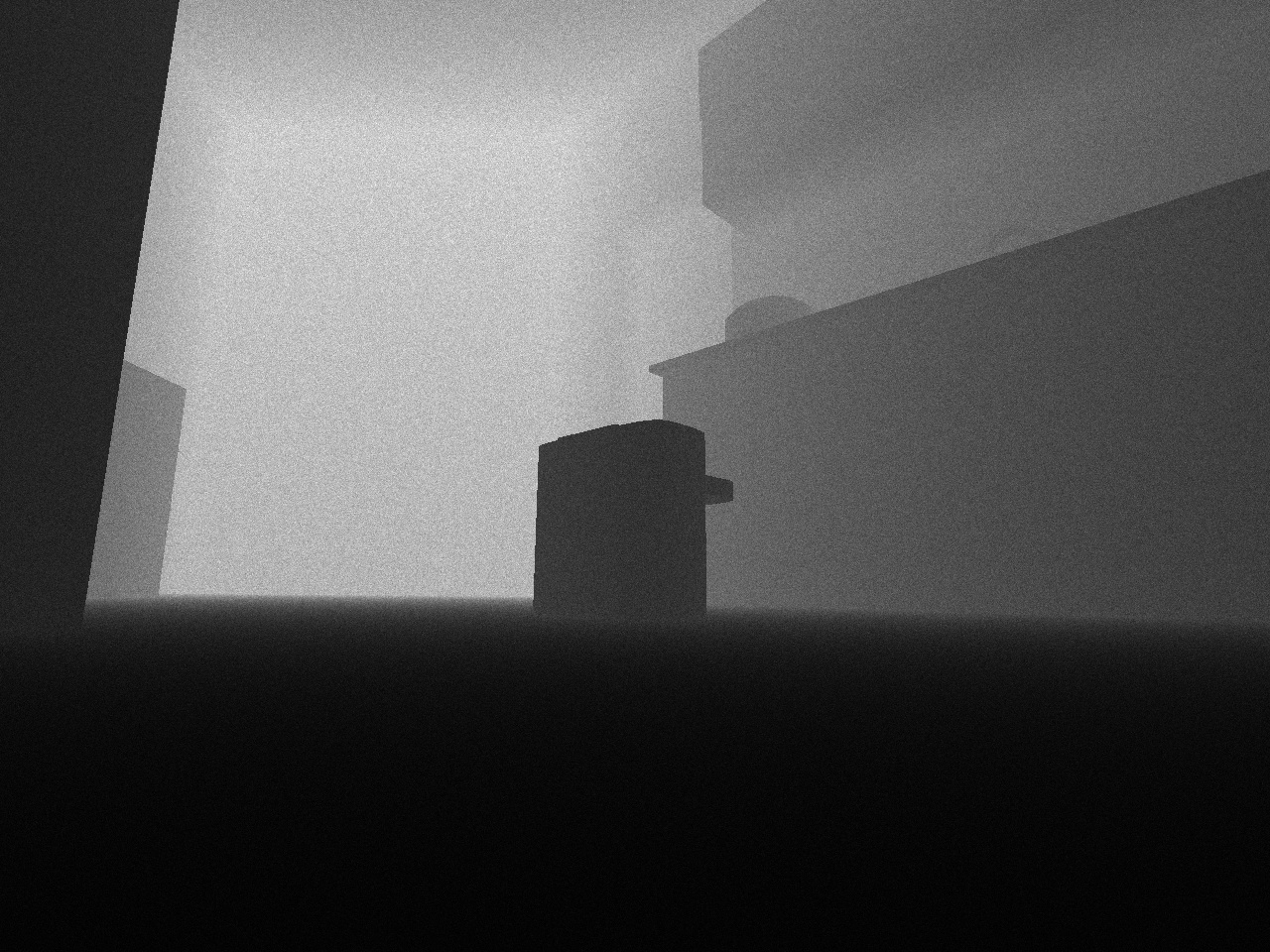} &
        \includegraphics[width=0.24\linewidth]{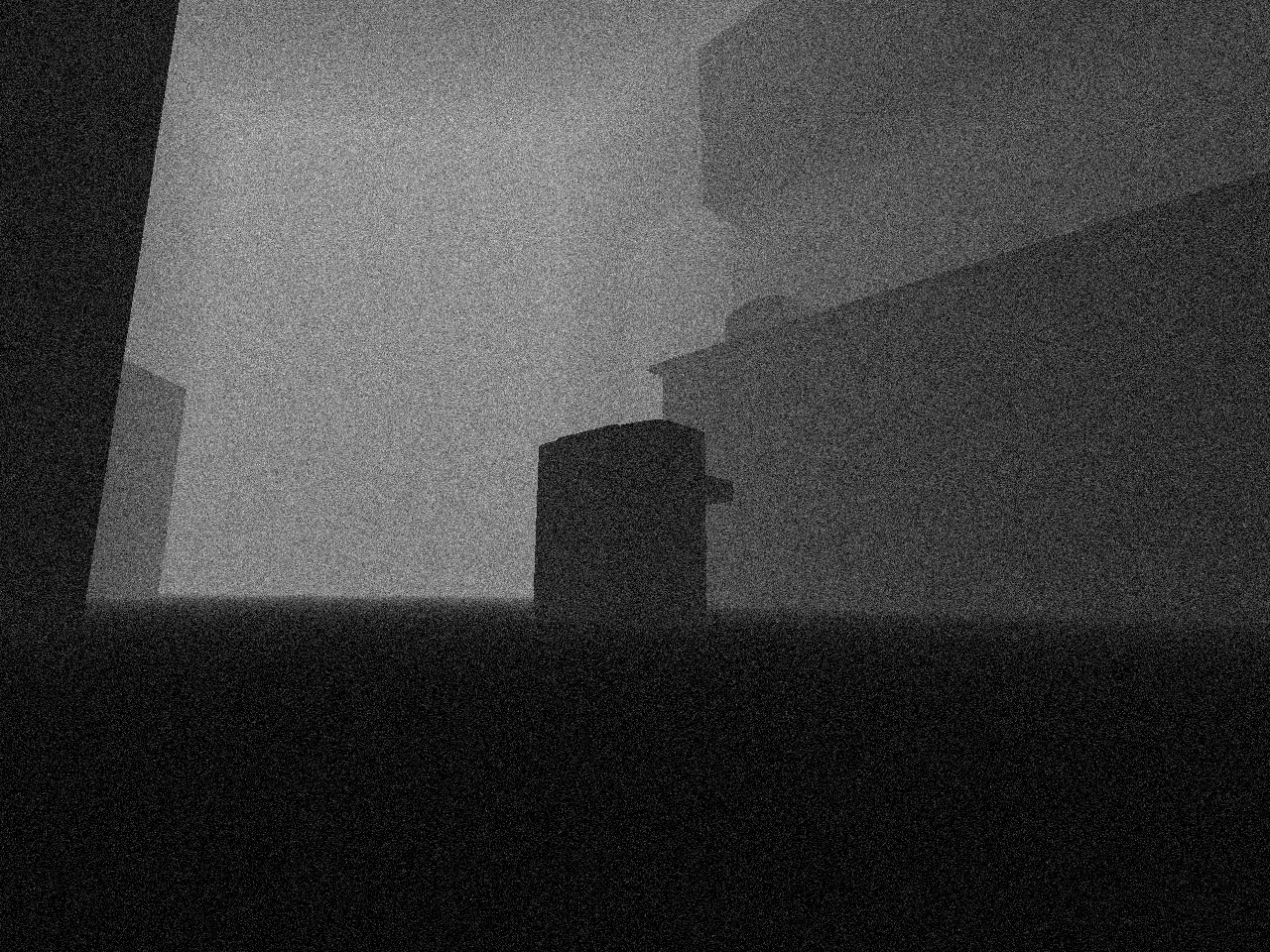} &
        \includegraphics[width=0.24\linewidth]{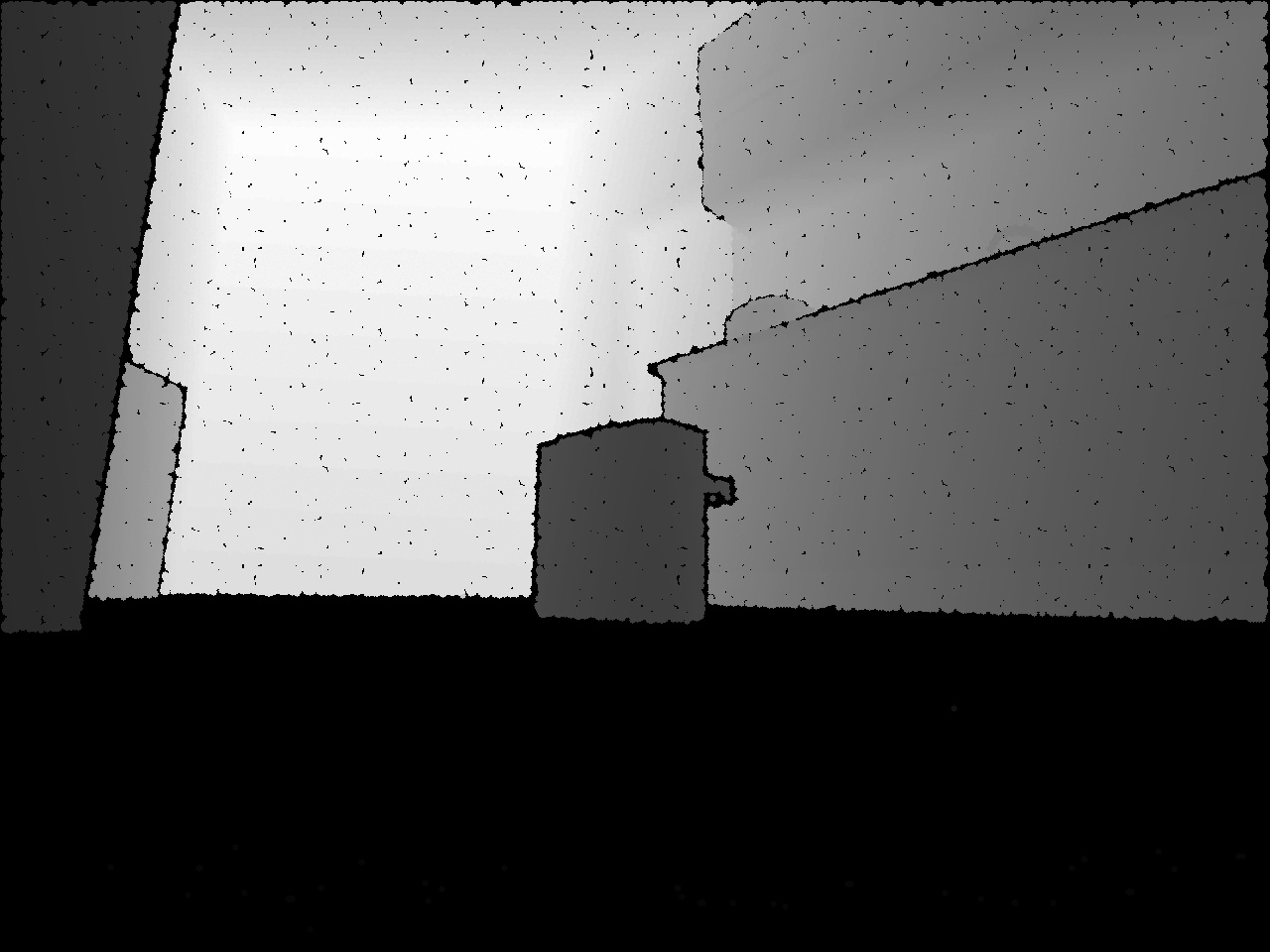} \\
        (a) & (b) & (c) & (d) \\
    \end{tabular}
    \caption{Results generated by the pixel-level sampler. (a) is the ground-truth depth output, whilst (b) add Poisson noise, (c) add Gaussian noise, and (d) add Gaussian shifts (Kinect noise).}
    \label{fig:PixelSampler}
\end{figure}

We provide several built-in functions, such as mapping the semantic labels to the desired label set, simulating different types of depth noise (e.g., Gaussian noise, Poisson noise, salt-and-pepper noise and Kinect noise~\cite{Barron:etal:2013A,handa:etal:2014,Bohg:etal:2014} as shown in Figure~\ref{fig:PixelSampler}). Users could also apply arbitrary customized image processing methods to the 2D renderings.

\section{DSL in MINERVAS}
\label{sec:dsl}

We design the DSL under two principles: flexibility and ease of use. For flexibility, we use the Python programming language as the host to build our internal DSL. Using its mechanism for assignment, arithmetic, loop, and function, users familiar with Python can easily manipulate data using our DSL. Users can also implement any algorithm for configuring and modifying scene attributes. For ease of use, we provide several built-in samplers for users to easily generate diverse scenes for domain randomization.

In this section, we show some use cases by the built-in functions and samplers of our system. Due to the space limitations, we refer readers to our project page for more examples.

\smallskip \noindent \textbf{Basic usage.} As our pipeline contains three different process stages, we provide corresponding processors, i.e., \mintinline{python}{SceneProcessor}, \mintinline{python}{EntityProcessor}, and \mintinline{python}{PixelProcessor} for users to manipulate data in each stage. To write DSL for a specific stage, the users need to declare a class inheriting from the corresponding built-in processor class. Then, the users need to write their customized operations in the \mintinline{python}{process} function.

\smallskip \noindent \textbf{Scene filtering.} In the Scene Process Stage, users could use DSL to filter scenes according to their requirements. The following code shows how to select the scenes with more than two rooms, and each room area should be larger than $20 \textrm{m}^2$:
\begin{pythoncode}
class SceneFilter(SceneProcessor):
    def process(self):
        if len(self.get_rooms) <= 2:
            self.filter_out()
        for room in self.get_rooms():
            if room.area < 20:
                self.filter_out()
\end{pythoncode}
We declare a class \mintinline{python}{SceneFilter} for the Scene Process Stage by inheriting the \mintinline{python}{SceneProcessor} class. Customized scene filtering operation are implemented in the \mintinline{python}{process} function.

\smallskip \noindent \textbf{Scene-level randomization.} Users could rearrange the furniture layout for each room using our built-in scene-level sampler:
\begin{pythoncode}
class FurnitureLayoutSampler(SceneProcessor):
    def process(self):
        for room in self.get_rooms():
            room.sample_layout()
\end{pythoncode}
Figure~\ref{fig:property}(a) shows the rearranged furniture layouts. Different from transforming each furniture randomly, our novel furniture layouts are reasonable because of considering multiple constraints from interior design guidelines.

\smallskip \noindent \textbf{Entity-level randomization.} Users could easily sample attributes of entities, such as material, CAD model, and light. In the following example, we randomly replace the CAD models and materials in the ``sofa'' and ``table'' categories and randomize lighting using the entity-level sampler:

\begin{pythoncode}
class EntitySampler(EntityProcessor):
    def process(self):
        for instance in self.shader.world.instances:
            if instance.label in [5, 8]:
                # 5: sofa, 8: table
                self.shader.world.replace_model(
                    id=instance.id)
                self.shader.world.replace_material(
                    id=instance.id)
        for light in self.shader.world.lights:
            light.tune_temp()
            light.tune_intensity()
\end{pythoncode}
Figure~\ref{fig:property} (b-d) shows the results of different entity-level samplers. With these samplers, the users can improve the diversity of the scene. Except for randomization, some fine-grained operations like the transformation of each furniture or light intensity of each light can also be manually configured.

\smallskip \noindent \textbf{Non-imagery data.} Our system also supports to export non-imagery data. We provide the function for users to export desired attributes easily. The following example shows the generation of camera information:
\begin{pythoncode}
class CustomUserOutput(EntityProcessor):
    def process(self):
        for camera in self.shader.world.cameras:
            # export the position and target of camera 
            self.shader.world.pick(
                type="camera", position=camera.position,
                target=camera.lookAt)
\end{pythoncode}
The \mintinline{python}{pick} method allows users to output any attributes as a JSON file. We utilize this feature to generate ground truths for the layout estimation task.

\smallskip \noindent \textbf{Pixel-level randomization.} The users could apply customized computations to the pixels. Here, we use the built-in pixel-level sampler to add different types of noise to the images:
\begin{pythoncode}
class DepthNoiseSampler(PixelProcessor):
    def process(self):
        # 4 represent Kinect noise model
        self.gen_depth(noise=4)
\end{pythoncode}
The results of multiple noise simulation are shown in Figure~\ref{fig:PixelSampler}.

\smallskip \noindent \textbf{Trajectory generation.} Finally, we show how to use our system to generate a video sequence by setting the camera trajectory:
\begin{pythoncode}
class TrajectorySampler(EntityProcessor):
    def process(self):
        for camera in self.shader.world.cameras:
            camera.set_attr("imageWidth", 640)
            camera.set_attr("imageHeight", 480)
            self.shader.world.add_trajectory(
                id='traj_'+camera.id, initCamera=camera,
                fps=3, speed=1200, time=5, height=1000,
                collisionPadding=300, type="RANDOM")
\end{pythoncode}
We first set the image resolution. Then, we add a trajectory to the camera with customized speed and frame rate. The sampler will generate a random trajectory using the built-in algorithm~\cite{handa2016scenenet}. Combining with scene layout information, the users can also generate camera trajectories by manually setting the key locations within the scene.

In our implementation, we employ the Entity Component System (ECS) architecture~\cite{garcia2014data} to organize the scenes for customization features in our system. As commonly used in 3D game engines, the ECS architecture provides a flexible and intuitive way to describe and manipulate scenes compared with the object-oriented design. The ECS architecture treats each object in the scene as an entity. Each entity contains various components storing attributes, e.g., transformation, CAD model, material. Our system decouples data from logic by utilizing the ECS architecture, which empowers users to plug in customized computations with DSL to control data synthesis procedures. Furthermore, we integrate the randomization feature into the ECS architecture by attaching a distribution to depict each component. The newly proposed architecture is named ECS-D, where D denotes distributions on components.

\begin{figure}[t]
	\small
    \centering
	\includegraphics[width=0.9\linewidth]{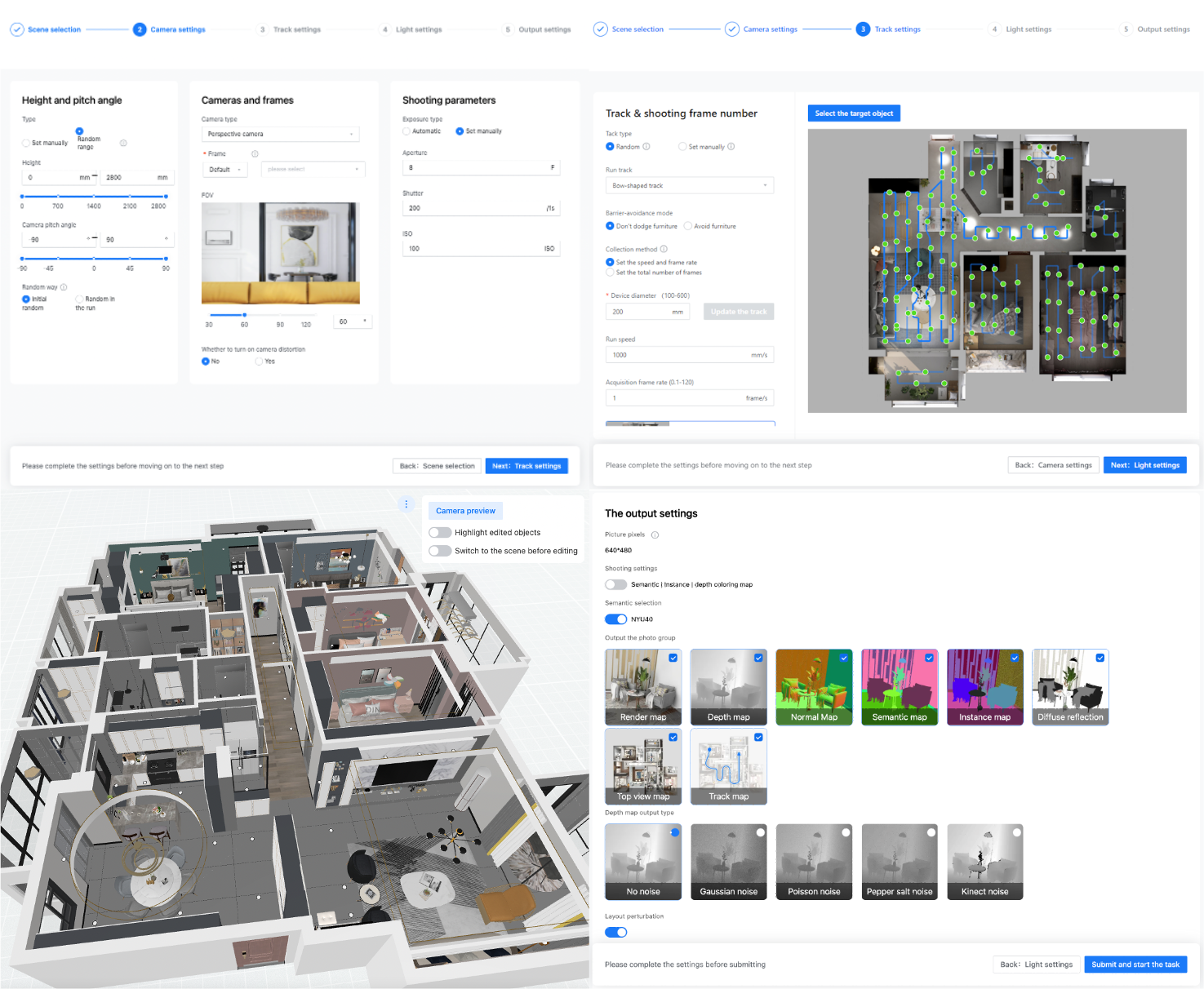}
	\caption{Screenshots of the GUI in our system. User-friendly GUI makes our system easier to use especially for novice users.}
	\label{fig:GUI}
\end{figure}

Based on the Python programming language, ECS-D architecture, and built-in samplers, different data syntheses can be customized by users via DSL easily. In our implementation, we also provide a user-friendly GUI to represent some functions in our DSL for novice users. Such as scene filtering by specific constraints, camera settings, trajectory editor, and renderer configurations. Users can also check each modified scene using the 3D viewer. Figure~\ref{fig:GUI} shows some screenshots of our GUI.

\section{Experiments}
\label{sec:experiments}

\begin{figure}[t]
	\small
    \centering
    \setlength{\tabcolsep}{1pt}
	\begin{tabular}{ccc}
		\includegraphics[width=0.32\linewidth]{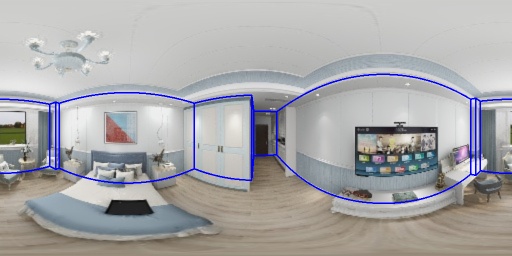} &
		\includegraphics[width=0.32\linewidth]{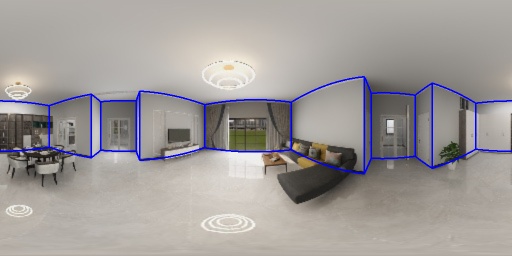} &
		\includegraphics[width=0.32\linewidth]{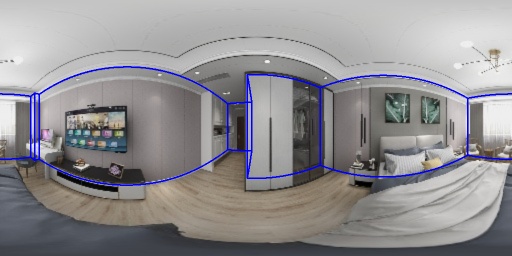} \\
	\end{tabular}
    \begin{tabular}{ccc}
        \includegraphics[width=0.32\linewidth]{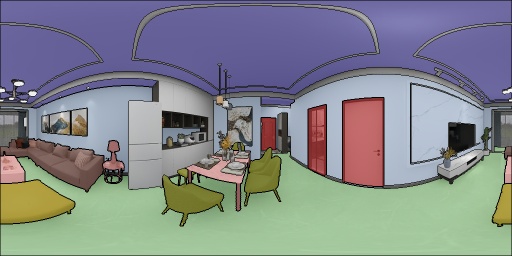} &
        \includegraphics[width=0.32\linewidth]{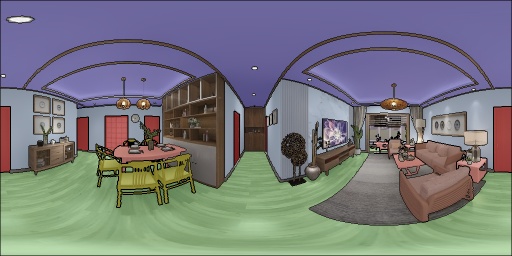} &
        \includegraphics[width=0.32\linewidth]{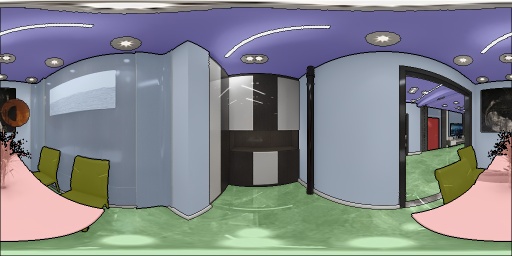} \\
    \end{tabular}
    \begin{tabular}{cccc}
		\includegraphics[width=0.24\linewidth]{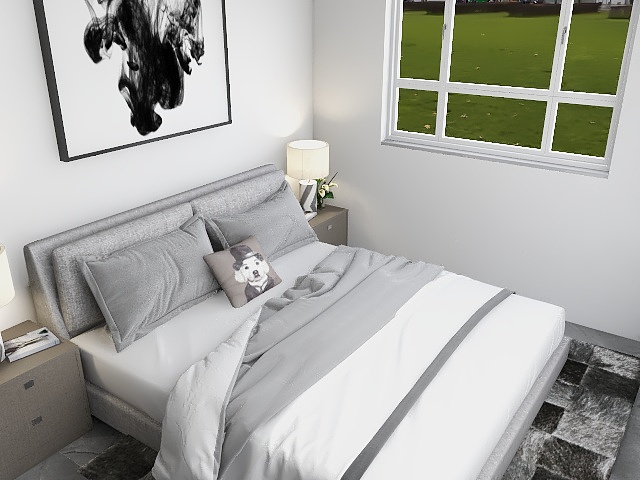} &
		\includegraphics[width=0.24\linewidth]{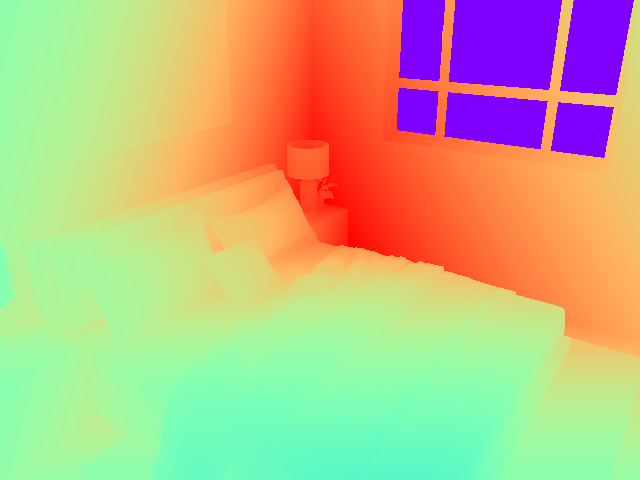} &
		\includegraphics[width=0.24\linewidth]{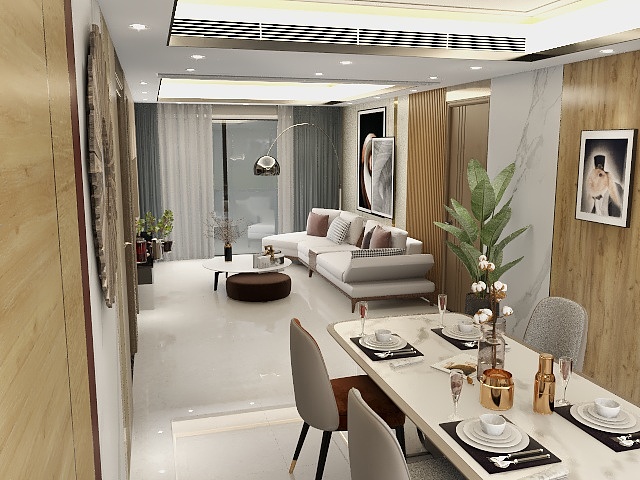} &
		\includegraphics[width=0.24\linewidth]{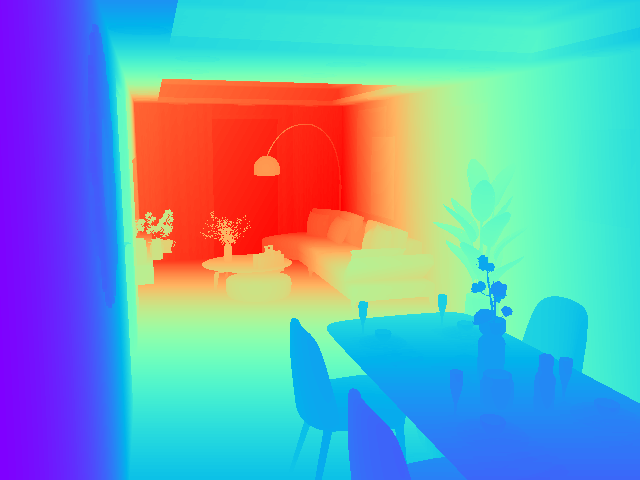} \\
    \end{tabular}
	\caption{Data samples generated by our system for Manhattan room layout estimation task (first row), semantic segmentation task (second row) and depth estimation task (third row).}
	\label{fig:ManhattanSamples}
\end{figure}

To verify the usefulness of our system, we use three crucial interior scene understanding tasks, including room layout estimation, semantic segmentation, and depth estimation. We use the DSL to customize the pipeline for data synthesis. We demonstrate the capability of our system from two following perspectives: (1) Flexibility: our system has the potential in different vision tasks. (2) Validity: our system can boost the performance of existing methods using the synthesized data.

For each task, the models are trained using two following training strategies: (1) training only on the real dataset (``r'') and (2) training on both synthetic and real dataset with Balanced Gradient Contribution (BGC)~\cite{RosSAW16} (``s + r''). All experiments are conducted on two NVIDIA Tesla V100 GPUs with 32 GB memory.

Due to the space limitations, we refer readers to the supplementary material for the DSL codes
and more quantitative and qualitative results.

\subsection{Manhattan Room Layout Estimation}
\label{sec:layout}

\begin{figure}[t]
	\small
    \centering
    \setlength{\tabcolsep}{1pt}
	\begin{tabular}{cc}
		\includegraphics[width=0.45\linewidth]{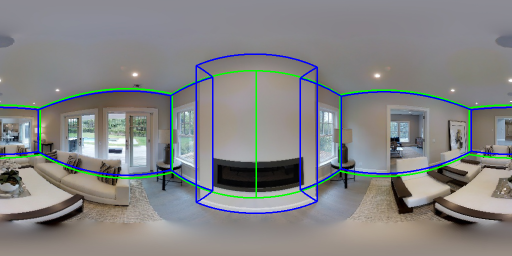} &
		\includegraphics[width=0.45\linewidth]{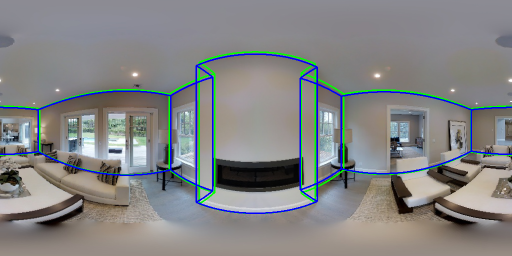} \\
		\includegraphics[width=0.45\linewidth]{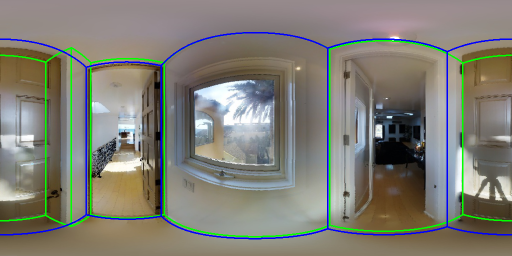} &
		\includegraphics[width=0.45\linewidth]{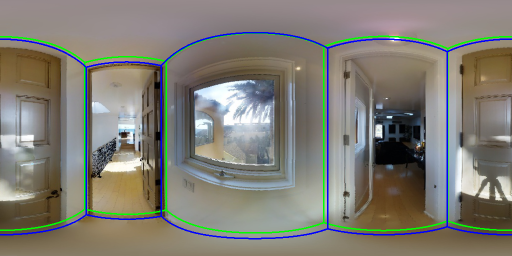} \\
		\includegraphics[width=0.45\linewidth]{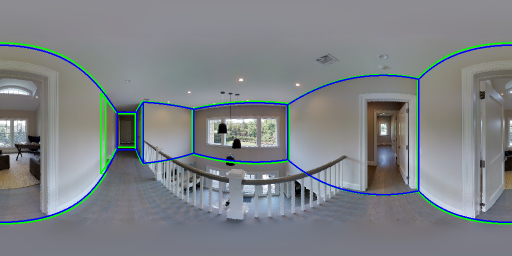} &
		\includegraphics[width=0.45\linewidth]{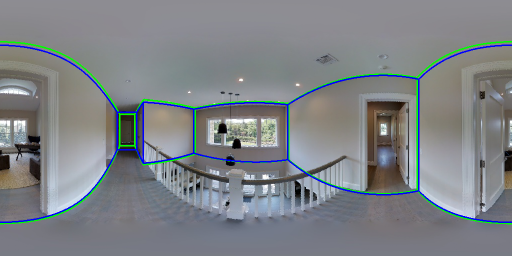} \\
		r & s + r\\
	\end{tabular}
	\caption{Qualitative results of the Manhattan room layout estimation on the MatterportLayout dataset. Blue lines denote the ground-truth layout, while green lines denote the predicted layout.}
	\label{fig:ManhattanResult}
\end{figure}

In this experiment, we first select scenes with Manhattan-world assumption in the Scene Process Stage. To increase the scene diversity, we also filter out the room with less than four furniture objects. We randomly place a panoramic camera with the resolution of $1024 \times 512$ in each room in the Entity Process Stage. Finally, we export the positions of room corners and camera parameters. We synthesize 120K panorama images using our system. We use MatterportLayout~\cite{chang2017matterport3d, ZouSPCSWCH21} as the real data. The MatterportLayout dataset consists of 1,647 images for training, 190 images for validation, and 458 images for testing.

We adopt HorizonNet~\cite{SunHSC19} as the baseline approach. We use an Adam optimizer with an initial learning rate of $3\times 10^{-4}$ with a polynomial decay policy. For ``r'', we set the mini-batch size to 24. For ``s + r'', each batch contains 16 images from the real dataset and 8 from the synthetic dataset. For each strategy, we train the network for 30K iterations.

\begin{table}[t]
    \footnotesize
    \centering
    \setlength{\tabcolsep}{5pt} 
    \caption{Manhattan room layout estimation results on MatterportLayout dataset.}
    \label{tab:layout}
    \begin{tabular}{lcccccc}
        \toprule
        Config. & \#Data & 3D IoU (\%) $\uparrow$ & 2D IoU (\%) $\uparrow$ & RMSE $\downarrow$ & $\delta_1$ $\uparrow$ \\
        \midrule
        r & - & 75.39 & 77.95 & 0.277 & 0.908 \\
        s + r & 1.2K & 76.56 & 79.22 & 0.275 & 0.912 \\
        s + r & 12K & 76.74 & 79.25 & 0.261 & \textbf{0.919} \\
        s + r & 120K & \textbf{76.92} & \textbf{79.49} & \textbf{0.258} & 0.918 \\
        \bottomrule
    \end{tabular}
\end{table}

We adopt four standard metrics: 3D IoU, 2D IoU, root mean squared error (RMSE) and threshold accuracy $\delta_1$ for evaluation. The results are shown in Table~\ref{tab:layout}. As can be seen, the model trained on both the synthetic and real datasets achieves better results. Furthermore, the model with more synthetic data is better.

\begin{table}[t]
    \footnotesize
    \centering
    \caption{Semantic segmentation results on 2D-3D-S dataset. DA: Domain Randomization. PA: Pixel Accuracy.}
    \label{tab:semantic}
    \begin{tabular}{lccccc}
        \toprule
        Config. & \#Scene & \#Data & DA & Mean IoU (\%) $\uparrow$ & PA (\%) $\uparrow$ \\
        \midrule
        r & - & - & - & 47.08 & 80.02 \\
        s + r & 6.7K & 30K & \xmark & 48.50 & 81.49 \\
        s + r & 26.7K & 120K & \xmark & \textbf{50.88} & 82.34 \\
        s + r & 6.7K & 120K & \cmark & 50.18 & \textbf{82.57} \\
        \bottomrule
    \end{tabular}
\end{table}

\subsection{Semantic Segmentation} 

\begin{figure}[t]
    \footnotesize
    \centering
    \setlength{\tabcolsep}{1pt}
    \begin{tabular}{ccc}
        \includegraphics[width=0.32\linewidth]{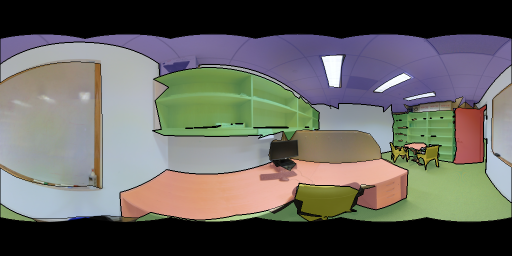} &
        \includegraphics[width=0.32\linewidth]{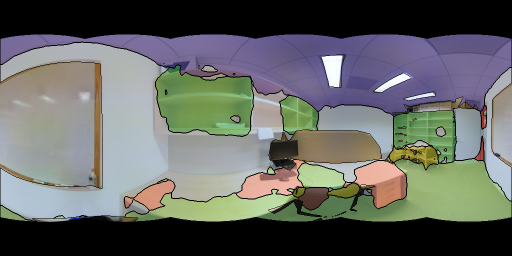} &
        \includegraphics[width=0.32\linewidth]{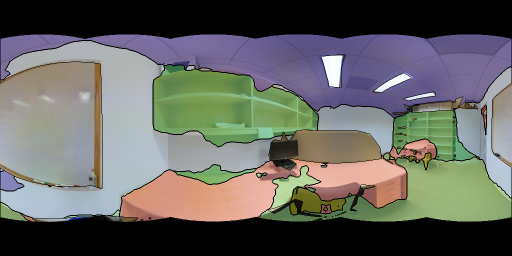} \\
        \includegraphics[width=0.32\linewidth]{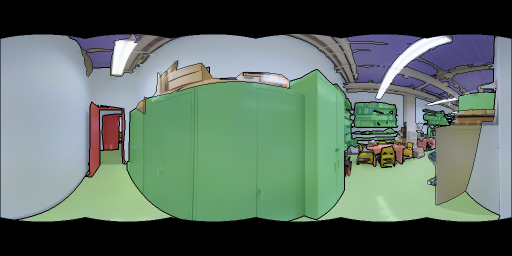} &
        \includegraphics[width=0.32\linewidth]{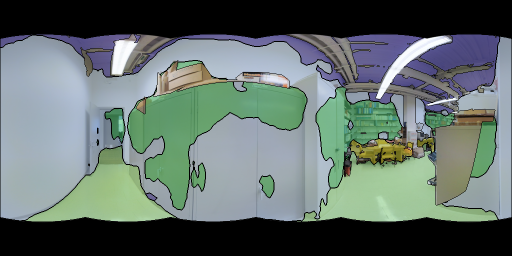} &
        \includegraphics[width=0.32\linewidth]{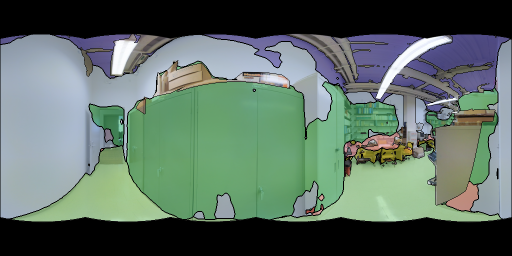} \\
        \includegraphics[width=0.32\linewidth]{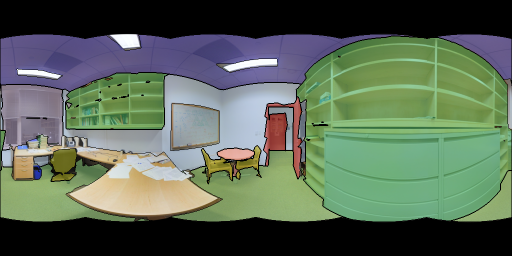} &
        \includegraphics[width=0.32\linewidth]{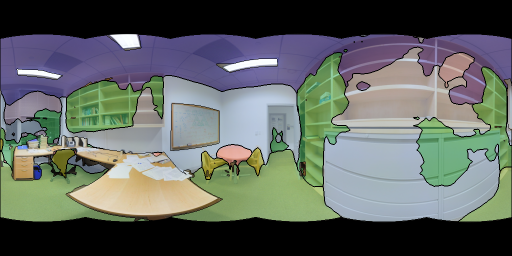} &
        \includegraphics[width=0.32\linewidth]{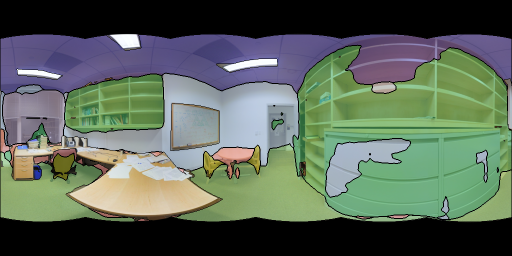} \\
        Ground truth & r & s + r \\
    \end{tabular}
    \caption{Semantic segmentation results of PSPNet. Different colors denote different semantic categories.}
    \label{fig:SemanticResult}
\end{figure}

In this experiment, we filter out the rooms with less than four furniture objects. In the Entity Process Stage, a panoramic camera with the $1024 \times 512$ resolution is randomly placed in the room. In the Pixel Process Stage, we export semantic labels with NYUv2 40 label set~\cite{Silberman:ECCV12}. We use our system to generate 120K panoramas. We use 1k images from 2D-3D-S~\cite{armeni2017joint} as the real data. Nine overlapping categories of the two datasets are used for evaluation: wall, floor, chair, sofa, door, window, bookshelf, ceiling, and table.

We use PSPNet~\cite{zhao2017pspnet} with dilated ResNet-50 as the backbone. We use an SGD optimizer with an initial learning rate of $2\times 10^{-2}$ with a polynomial decay policy, momentum 0.9, and weight decay of $10^{-4}$. We set the mini-batch size to $8$. In ``s + r'', each batch contains 4 images from the real dataset and 4 from the synthetic dataset. For each strategy, we train the whole network for 20K iterations.

We adopt two standard metrics: Mean IoU and Pixel Accuracy (PA). The results are reported in Table~\ref{tab:semantic}. Training with our synthetic dataset achieves better performance. The performance can be further improved when using more scenes. Furthermore, we conduct an ablation study on Domain Randomization (DR) of layout, camera, light, model, and material. We generate more images (120K) with limited scenes (6.7K) by incorporating DR technique. When using the same number of scenes, the result with DR (the fourth row) is better than that without DR (the second row). The result with generated scenes by DR is comparable with that using more original designed scenes (the third row). This verifies the effectiveness of DR.

\begin{table}[t]
    \footnotesize
    \centering
    \setlength{\tabcolsep}{1pt}
    \caption{Depth estimation results on NYUv2 dataset.}
    \label{tab:depth}
    \begin{tabular}{lccccccc}
        \toprule
        Methods & Config. & Rel $\downarrow$ & $\log_{10} \downarrow$ & RMSE $\downarrow$ & $\delta_1 \uparrow$ & $\delta_2 \uparrow$ & $\delta_3 \uparrow$ \\
        \midrule
        \multirow{2}{*}{VNL~\cite{yin2019enforcing}} & r & 0.133 & 0.057 & 0.573 & 0.833 & 0.969 & 0.991 \\
        & s + r & \textbf{0.116} & \textbf{0.050} & \textbf{0.528} & \textbf{0.869} & \textbf{0.976} & \textbf{0.994} \\
        \midrule
        \multirow{2}{*}{AdaBins~\cite{bhat2021adabins}} & r & 0.134 & 0.055 & 0.452 & 0.841 & 0.972 & 0.992 \\
        & s + r & \textbf{0.129} & \textbf{0.053} & \textbf{0.421} & \textbf{0.856} & \textbf{0.975} & \textbf{0.995} \\
        \bottomrule
    \end{tabular}
\end{table}

\subsection{Depth Estimation} 
\label{sec:depth}

\begin{figure}[t]
    \footnotesize
    \centering
    \setlength{\tabcolsep}{1pt}
    \begin{tabular}{cccc}
        \includegraphics[width=0.24\linewidth]{Images/Supp/DepthEstimation/VNL/rgb/nyudv2_0316_r.png} &
        \includegraphics[width=0.24\linewidth]{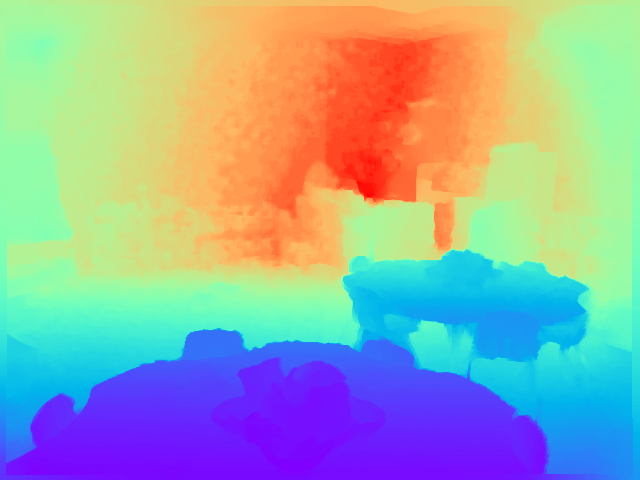} &
        \includegraphics[width=0.24\linewidth]{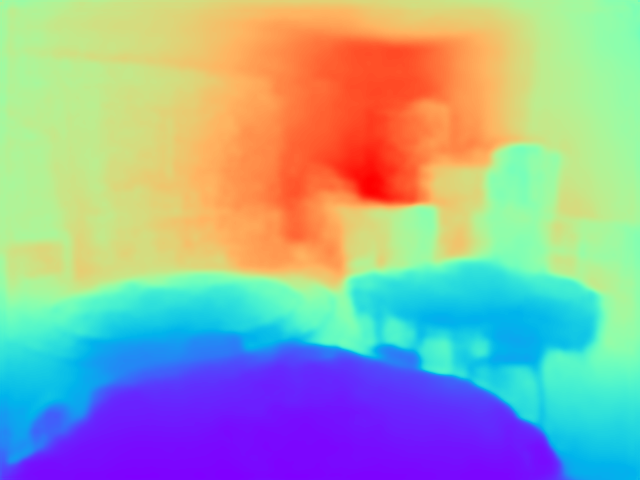} &
        \includegraphics[width=0.24\linewidth]{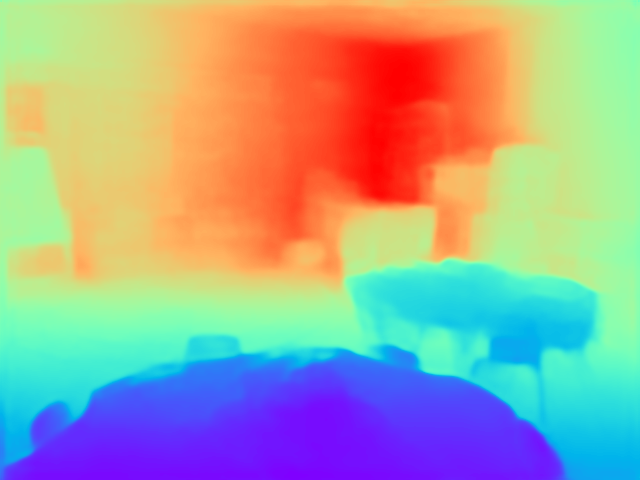} \\
        \includegraphics[width=0.24\linewidth]{Images/Supp/DepthEstimation/VNL/rgb/nyudv2_1207_r.png} & 
        \includegraphics[width=0.24\linewidth]{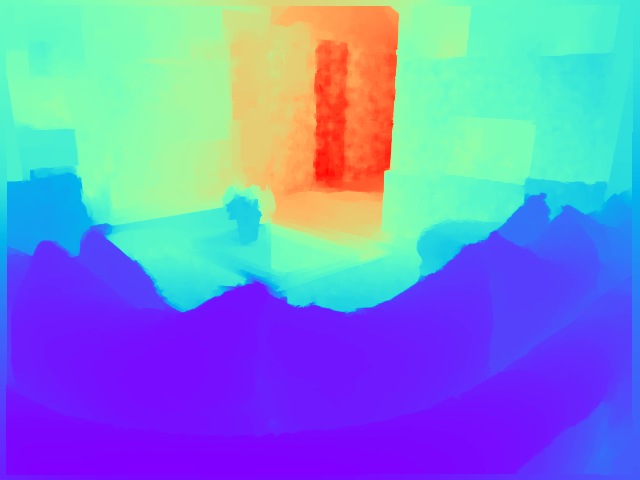} & 
        \includegraphics[width=0.24\linewidth]{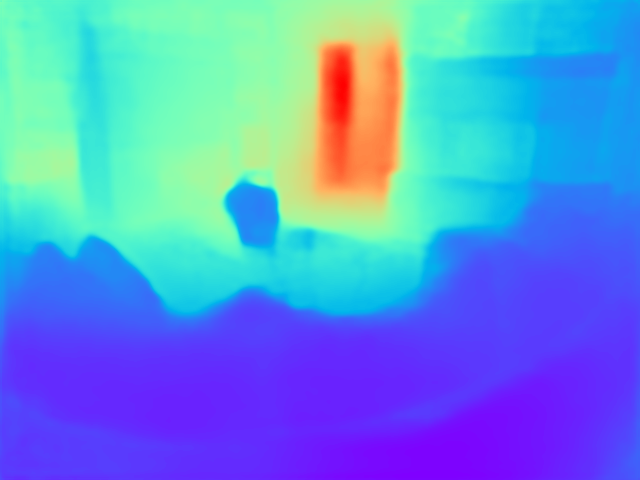} & 
        \includegraphics[width=0.24\linewidth]{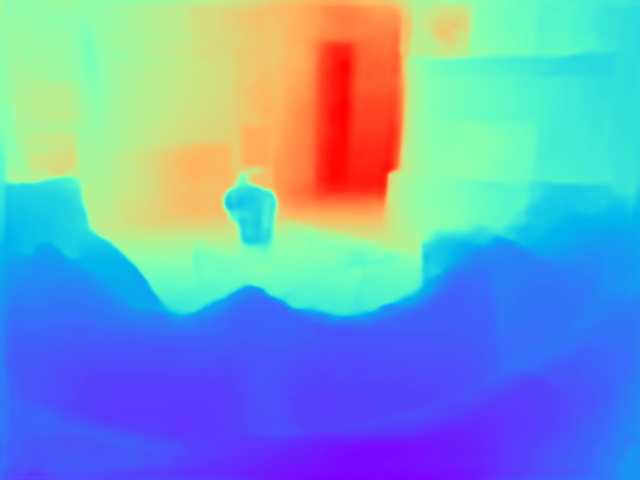} \\
		\includegraphics[width=0.24\linewidth]{Images/Supp/DepthEstimation/VNL/rgb/nyudv2_1411_r.png} &
		\includegraphics[width=0.24\linewidth]{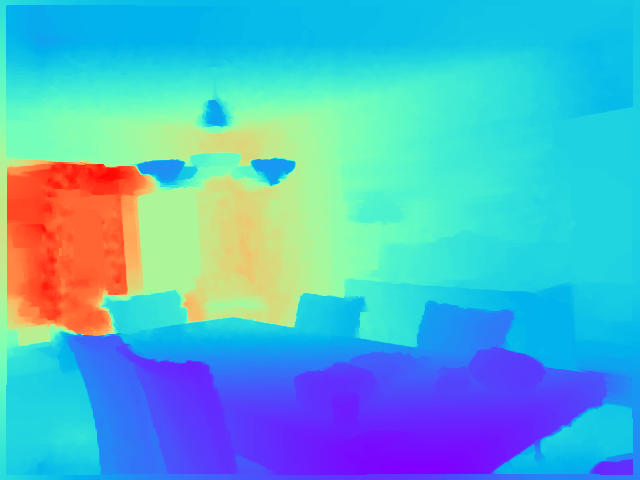} &
		\includegraphics[width=0.24\linewidth]{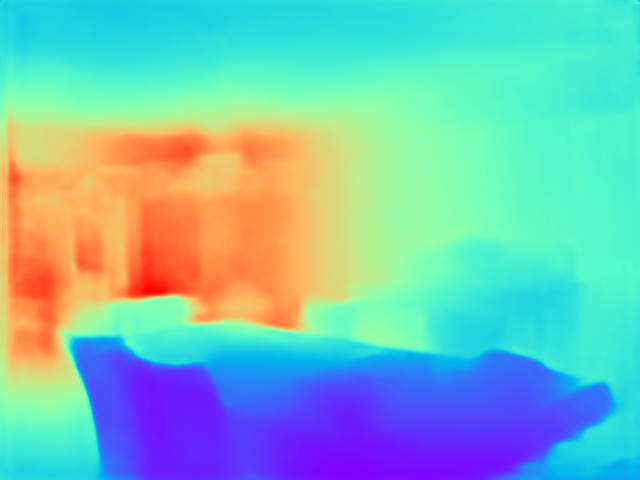} &
		\includegraphics[width=0.24\linewidth]{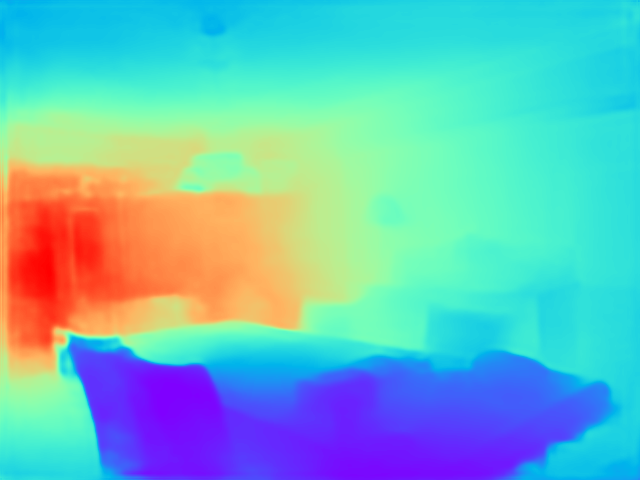} \\
        RGB & Ground truth & r & s + r \\
    \end{tabular}
    \caption{Qualitative results of the depth estimation with VNL~\cite{yin2019enforcing}.}
    \label{fig:DepthResult}
\end{figure}

In this experiment, the camera are placed by customized heuristic rules.
We set the image resolution as $640 \times 480$ and the horizontal field-of-view (FoV) to $57$ degree in the Entity Process Stage. The depth images are exported in the Pixel Process Stage. We use our system to generate 120k data. We use 1449 images from labeled NYUv2 dataset~\cite{Silberman:ECCV12} as the real data. 

We adopt two methods for this task: VNL~\cite{yin2019enforcing} and AdaBins~\cite{bhat2021adabins}. For VNL~\cite{yin2019enforcing}, we use an SGD optimizer with an initial learning rate $1\times 10^{-4}$ with polynomial decay policy, momentum $0.9$, and weight decay $5 \times 10^{-4}$. We set the mini-batch size to $8$. In ``s + r'', each batch contains 4 images from the real dataset and 4 from the synthetic dataset.
For AdaBins~\cite{bhat2021adabins}, we use an AdamW optimizer with max learning rate $0.000357$ with 1-cycle learning rate decay policy~\cite{smith2019super}, and weight decay $0.1$. We set the mini-batch size to $16$. In ``s + r'', each batch contains 10 images from the real dataset and 6 from the synthetic dataset. For each strategy in both methods, we train the whole network for 40K iterations.

To perform a reasonable evaluation, we use four common metrics: average absolute relative error (Rel), average $\log_{10}$ error ($\log_{10}$), room mean squared error (RMSE) and threshold accuracy ($\delta_i < 1.25^i$, where $i = 1, 2, 3$). Table~\ref{tab:depth} summarizes the quantitative results. It can be seen that augmenting real datasets with our synthetic data achieves better performances for both methods.

\smallskip \noindent \textbf{Data generalization.} We further conduct an experiment to compare the generalization of the models trained on our synthetic dataset or the real dataset. We use an accurate and diverse real-world captured depth dataset, DIODE dataset~\cite{diode_dataset}, as the test dataset. We use the NYUv2 dataset~\cite{Silberman:ECCV12} and our synthetic dataset as training data separately. In this experiment, we use AdaBins~\cite{bhat2021adabins} as the baseline approach.

The results are shown in Table~\ref{tab:diode}. As one can see, the model trained on our synthetic data has better generalization performance than that trained on NYUv2~\cite{Silberman:ECCV12}, even with the same number of training samples. This may be due to the limited depth ranges in the NYUv2 dataset. In contrast, our dataset and DIODE dataset has more diverse depth samples. When using more training data from our dataset, the performance is significantly improved.

\begin{table}[t]
    \footnotesize
    \centering
    \setlength{\tabcolsep}{3pt}
    \caption{Data generalization results on the DIODE dataset~\cite{diode_dataset}. The best and the second best results are boldfaced and underlined.}
    \label{tab:diode}
    \begin{tabular}{lccccccc}
        \toprule
        Train Set & \#Data &  Rel $\downarrow$ & $\log_{10} \downarrow$ & RMSE $\downarrow$ & $\delta_1 \uparrow$ & $\delta_2 \uparrow$ & $\delta_3 \uparrow$ \\
        \midrule
        NYUv2 & 795 & \textbf{0.463} & 0.297 & 2.600 & 0.181 & 0.409 & 0.605 \\
        Ours & 795 & 0.509 & \underline{0.255} & \textbf{2.386} & \underline{0.283} & \underline{0.517} & \underline{0.666} \\
        Ours & 120K & \underline{0.503} & \textbf{0.249} & \underline{2.513} & \textbf{0.359} & \textbf{0.580} & \textbf{0.712} \\
        \bottomrule
    \end{tabular}
\end{table}

\section{Conclusion}
\label{sec:conclusion}

This paper presents MINERVAS, a programmable system for interior imagery data synthesis. The system employs ECS-D architecture for scene representation and introduces an easy-to-use and flexible DSL for task-specific customization. Additionally, we introduce multi-level samplers for randomization to increase the diversity of the synthetic data. The experimental results validate the ability of our system to boost the performance on various interior scene understanding tasks.

\smallskip \noindent \textbf{Limitations and future work.} 
In order to protect the copyright of 3D assets, explicit 3D mesh data can not be obtained from our system. Thus our system can not facilitate some tasks operating on the mesh directly. As our system can generate any geometry cues in the image space, our system can still facilitate a broad range of applications in scene understanding or scene synthesis. Since we mainly focus on computer vision tasks, our system currently does not support physics and interaction with virtual scenes. In the future, we plan to integrate the physics and real-time interactive simulation into the MINERVAS system for online embodied agent learning. Another promising direction for future work would be how to automatically synthesize data to minimize domain gaps between synthetic data and real-world data.
Though our system protects 3D assets via remote rendering, the users might illegally reconstruct the 3D models using recent 3D reconstruction techniques. It is important to incorporate some advanced techniques of preventing reconstruction~\cite{koller2004protected, koller2005protecting} in the future. There may exist several ways to further protect the copyright in our system. First, we can detect suspicious sequences or frequencies of image requests~\cite{koller2005protecting} in the system. Secondly, some watermarks (e.g., subtle noise) can be added to rendered images to prevent the accurate reconstruction~\cite{koller2005protecting}. The adversarial attack technique~\cite{goodfellow2014explaining} may also help this. Finally, the blockchain technique~\cite{park2021meshchain} may also be applied to help identify the illegal release to protect the copyright.

\section*{Acknowledgments}

This work was supported in part by NSFC (No. 61872319), Key R\&D Program of Zhejiang Province (No. 2022C01025), and the Fundamental Research Funds for the Central Universities. We would like to thank Manycore Tech Inc. (Coohom \& Kujiale) for providing the 3D indoor scene database and the high-performance rendering platform, especially Coohom Cloud Team for technical support.

\bibliographystyle{eg-alpha-doi} 
\bibliography{paper}

\end{document}